%% file: sample_paper.tex
\documentclass[twoside]{article}
\usepackage[accepted]{aistats2021}
\usepackage[round]{natbib}

\input{definitions}

\begin{document}

% If your paper is accepted and the title of your paper is very long,
% the style will print as headings an error message. Use the following
% command to supply a shorter title of your paper so that it can be
% used as headings.
%
\runningtitle{Offline detection of change-points in the mean for stationary graph signals}

% If your paper is accepted and the number of authors is large, the
% style will print as headings an error message. Use the following
% command to supply a shorter version of the authors names so that
% they can be used as headings (for example, use only the surnames)
%
%\runningauthor{Surname 1, Surname 2, Surname 3, ...., Surname n}

\newcommand{\Dtau}{d}%{D_{\tau}}

\twocolumn[

\aistatstitle{Offline detection of change-points in the mean\\ for stationary graph signals}
%\aistatstitle{Offline change-point detection in the mean of stationary graph signals}

\aistatsauthor{ Alejandro de la Concha \And Nicolas Vayatis \And Argyris Kalogeratos }

\aistatsaddress{ Université  Paris-Saclay, ENS Paris-Saclay, CNRS,  Centre Borelli, %F-91190 Gif-sur-Yvette, 
France 
} ]

\begin{abstract}
This paper addresses the problem of \emph{segmenting a stream of graph signals}: we aim to detect changes in the mean of a multivariate signal defined over the nodes of a known graph. We propose an offline method that relies on the concept of \emph{graph signal stationarity}  and allows the convenient translation of the problem from the original vertex domain to the spectral domain (Graph Fourier Transform), where it is much easier to solve. Although the obtained spectral representation is sparse in real applications, to the best of our knowledge this property has not been sufficiently exploited in the existing related literature. Our change-point detection method adopts a model selection approach that takes into account the sparsity of the spectral representation and determines automatically the number of change-points. Our detector comes with a proof of a non-asymptotic oracle inequality. Numerical experiments demonstrate the performance of the proposed method.
\end{abstract}

\section{Introduction}
\label{sec:intro}

One of the most common tasks in Signal Processing is segmentation. Identifying time intervals where a signal is `homogeneous' is a way to uncover latent features of its source. 
The signal segmentation problem can be restated as a change-point detection task: delimiting a segment can be achieved by spotting the timestamps where it starts and ends. This subject has being extensively investigated leading to a vast literature and applications in many domains including computer science, finance, medicine, geology, meteorology, \etc The majority of the related work so far %in signal segmentation 
focuses on temporal signals \citep{Baseville1993,Balzano2010,Chen2012,Tartakovsky2014,Aminikhanghahi2016,truong2020}.

In this work we study a different type of object: \emph{graph signals appearing as a stream}. In general terms, a graph signal is a function defined over the nodes of a given graph. Intuitively, the graph partially encodes the variability of the function, \ie nodes that are connected will take similar values. This applies to real situations, \eg contacts in social networks would share similar tastes, two neighboring sensors of a sensor network would provide similar measurements, etc. %\NOTE{%Moreover, this property is not only present when the graph is explicitly given.
Note that this property can be exploited even when the graph is not %explicitly 
given. 
%} 
In some applications, the graph itself has to be inferred and most algorithms are built over this property of local similarity of nodes' behavior that corresponds to signal \emph{smoothness} \citep{Kalofolias16,LeBarsICASSP2019}. This can be seen also in graphical models or graphs used to approximate manifolds \citep{Perraudin2017,Friedman2007,LeBars2020icml,Tenenbaum2000}. 

Despite the plethora of change-point detection methods in the literature, the development of detectors specifically for graph signals is still limited \citep{Balzano2010,Angelosante2011,Chen2018,LeBars2020icml}. %To the best of our knowledge, 
Most existing methods do not take into account the interplay between the signal and the graph structure. The main contribution of this article is in exactly this direction: an offline change-point detector aiming to spot jumps in the mean of a Stream of Graph Signals (SGS). Our method leverages many of the techniques developed in Graph Signal Processing% %(GSP)
, a relatively new field aiming to generalize the tools commonly used in classical Signal Processing \citep{Shuman2013,Ortega2018}. More specifically, our approach depends on the concept of Graph Fourier Transform (GFT) that, similarly to the usual Fourier Transform, induces a spectral domain and a sparse representation of the signal. The main idea behind our approach is to translate the problem from the vertex domain to the spectral domain, and design a change-point detector operating in this space that accounts for the sparsity of the data and automatically infers the number of change-points. This is achieved by adding two penalization terms:
one aiming to recover the sparsity, and another one penalizing models with a large number of change-points. The performance of the algorithm and the design of this penalization terms are based on the framework introduced in \cite{Birge2001} and the innovative perspective of the $\ell_1$ norm analyzed in \cite{Massart2011}.

The paper is organized as follows. \Sec{sec:basic_definitions} presents the basic definitions and tools that are used later on. \Sec{sec:our_method} formulates the change-point detection problem in the context of graph signals, and then presents a Lasso-based algorithm and a Variable Selection-based algorithm. \Sec{sec:Model_selection} provides theoretical guarantees for these algorithms and, finally, in \Sec{sec:exps} our overall approach is tested through experiments on synthetic signals generated over real and random graphs.

\section{Basic concepts and notations}{\label{sec:basic_definitions}}

In this section we introduce notations and key concepts.
Let $A_i$ and $A^{(j)}$ denote the $i$-th row and $j$-th column of matrix $A$, respectively, and $A_{i,j}$ be a specific element in $A$. Moreover, $A^{\Top}$ and $A^*$ stand for the transpose and the conjugate transpose (\ie transpose with negative imaginary part) of matrix $A$. $x^{(i)}$ denotes the $i$-th entry of vector $x$, $x_t$ represents the observed vector $x$ at time $t$, and $\tilde{x}$ stands for the GFT of $x$, which is introduced in 
\Definition{GFT}. A graph is defined by an ordered  tuple $G=(V,E)$, where $V$ and $E$ stand for the set of vertices and edges respectively, and $p := |V|$ is the number of graph nodes.

\begin{definition}{\label{Graph Signal}}
A \textbf{graph signal} is a tuple $(G,y)$, where $G=(V,E)$  and $y$ is a function $y:V \rightarrow \mathbb{R}$.
\end{definition} 

\begin{definition}{\label{GSO}}
A \textbf{graph shift operator} (GSO) $S$ associated with a graph $G=(V,E)$, is a $p \times p$ matrix whose entry $S_{i,j} \neq 0 $ iff $i=j$ or $(i,j)\in E$, and it admits an eigenvector decomposition $S=U\Theta U^*$.
\end{definition}

\begin{definition}{\label{GFT}}
For a given GSO $S=U\Theta U^*$ associated with a graph $G$, the \textbf{Graph Fourier Transform} (GFT) of a graph signal $y: V \rightarrow \R$ is defined by $\tilde{y}= U^*y$. 
\end{definition}

The frequencies of the GFT  correspond to the elements of the diagonal matrix $\Theta$, that is $\{\Theta_{i,i}\}_{i=1}^p$. Moreover, the eigenvectors $\{u_i\}_{i=1}^p$ provide an orthogonal basis for the graph signals defined over the graph $G$. Finally, the GFT is the basic tool that allows us to translate operations from the vertex domain to the spectral domain \citep{Sandryhaila2013}.
 
The \emph{graph signal stationarity} over the vertex domain is a property aiming to formalize the notion that the graph structure can explain to a large degree the observed inter-dependencies between the dimensions of a graph signal. Henceforth, we refer to stationary with respect to a GSO $S$. The definitions and the properties that are listed bellow can be found in \cite{Marques2017} and in \cite{Perraudin2017}. 

\begin{definition}{\label{Stationarity2}}
\textbf{Stationarity \wrt the vertex domain:}
Given a normal GSO $S$, a zero-mean graph signal $y:V \rightarrow R$ with covariance matrix $\Sigma_y$ is stationary with respect to the vertex domain encoded by $S$, iff $\Sigma_y$ and $S$ are simultaneously diagonalizable, \ie $\Sigma_y=U \diag(P_y) U^{\Top}$. The vector $P_y \in \R^p$ is known as the \textbf{graph power spectral density} (PSD).
\end{definition}

The next two properties are used to derive our change-point detection algorithm, to perform the estimation of $P_y$, and to generate synthetic experimental scenarios.

\begin{Prop}{\label{Property:independence}}
Let $y$ be a stationary graph signal \wrt to $S$, then $\tilde{y}=U^* y$, which means that the GFT of $y$ will have a covariance matrix $\Sigma_{\tilde{y}}= \diag (P_y)$.
\end{Prop}
\begin{Prop}{\label{prop:GF_stationarity}}
Let $y$ be a stationary graph signal with covariance matrix $\Sigma_y$ and PSD $P_y$. The output of a graph filter $H=U \diag(h) U^*$, with a frequency response $h$, applied to the graph signal is $z = Hy$ and has the following properties:
%
%\vspace{-0.7em}
%\begin{enumerate}\itemsep0.5em\leftmargin0.0em\topmargin0.5em
%    \item It's stationary on $S$ with covariance $\Sigma_z=H \Sigma_y H^*$.
%    \item  $P_z^{(i)}= |h^{(i)}|^2 P_y^{(i)}$.
%\end{enumerate}
\\
\noindent  1. It's stationary on $S$ with covariance $\Sigma_z=H \Sigma_y H^*$.\\
\noindent 2. $P_z^{(i)}= |h^{(i)}|^2 P_y^{(i)}$.\\
\end{Prop}

\section{Change-point detection for a stream of graph signals}\label{sec:our_method}

\subsection{Problem formulation}

\begin{figure}[t]\centering
\begin{minipage}[t]{1\linewidth}\centering
   \includegraphics[width=\linewidth,viewport=50 40 380 260,clip]{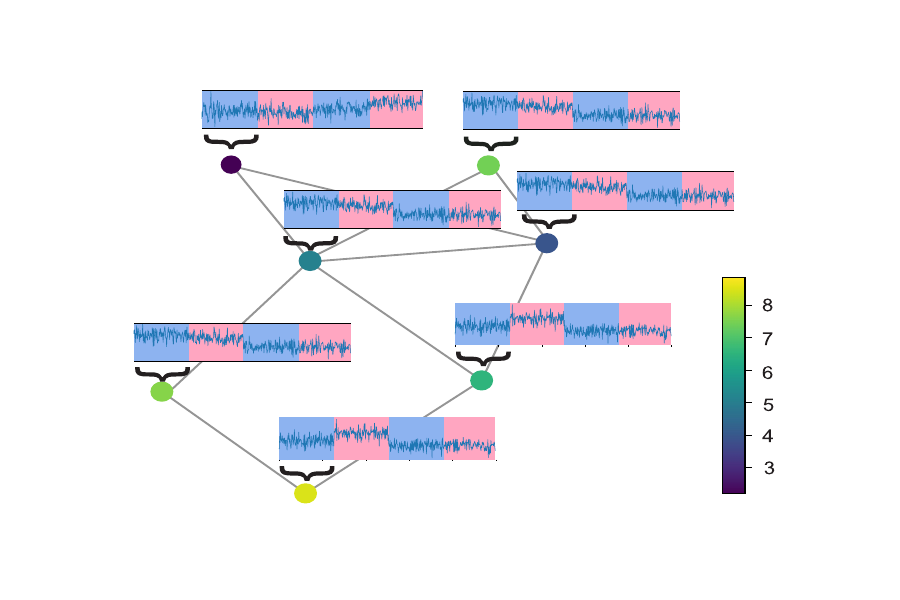}%
\end{minipage}%
\vspace{-0.5em}
\caption{ \small{Example stream of graph signals (SGS) with five change-points in the mean (according our problem formulation, %we take into account 
the end of the sequence is always a change-point). Successive segments are shown with different colors. The color of the graph nodes represents the mean of the signal during the first observed segment% of the observed signal
. The signal observed 
at each node evolves through time, as shown in the line plots next to them. At some timestamps, the mean of the graph signals exhibits a change in a subset of nodes, which also signifies changes in the spectral representation of the signals.}
}\vspace{-0.5em}
\label{fig:CP_GS}
\end{figure}

Suppose a stream of graph signals (SGS) $Y$ observed over the same graph $G=(E,V,S)$. Let $Y=\{y_t\}_{t=0}^{T}$,
where $\forall t\!: y_t \in \R^{p}$, and also $\mu_t = \Expec{y_t}$ is its unknown mean value. These expected values are the rows of matrix $\mu \in \R\ \!\!\!\!^{T \times p}$. We suppose that there is an unknown ordered set $\tau=\{\tau_0=0,...,\tau_{\Dtau}=T\} \subset \{0,...,T\}$ indicating $\Dtau+1$ \emph{change-points}, which segments the time-series. Our hypothesis is that the expected values in each of the segments induced by $\tau$ remains constant. Our goal is to infer the set of change-points $\tau$ and the $\mu$ that will be an element of the space: %
\begin{equation}
%\R^{T \times p}
\!\!\!\!\!F_{\tau} = \{\mu \in \R\ \!\!\!\!^{T \times p} \ |\ \mu_{\tau_{l-1}+1}=...=\mu_{\tau_{l}}, \forall \tau_{l} \in \tau\backslash \{0\}\}.\!\!
\end{equation}
The problem is illustrated in \Fig{fig:CP_GS} through an example where we can identify $4$ different segments, \ie $\Dtau+1=|\tau|=5$ change-points. 

The general idea of our approach is to take advantage of the advances in Graph Signal Processing and model selection fields in order to recover automatically the number of change-points as well as sparse representations of the $\{\mu_{\tau_l}\}_{l \in \{1,...,\Dtau\}}$ in terms of the GFT. This allows not only to segment the signal, but also to  characterize the nature of the change. Such an example is later illustrated in \Fig{fig:CP_Manhatan}.% illustrates an experiment carried out over the nodes of the graph induced by the intersections of the streets at Minnesota.  

We make the following hypotheses over the SGS:
\vspace{-1em}
%\inlinetitle{Hypotheses}{:}
\begin{enumerate}[leftmargin=1.7em]\itemsep-0.2em 
    \item The graph signals are independent with respect to the temporal domain.
    \item The graph signals follow a multivariate Gaussian distribution sharing the same expectation parameter if they belong to the same segment.
    \item If $t \in \{\tau_{l-1}+1,\tau_{l-1}+2,...,\tau_{l}\}$, then the residual $y_t-\mu_{\tau_l}$ is stationary with respect to the GSO $S$. 
    \item The mean of the graph signals admits a sparse representation with respect to the basis defined by the eigenvectors of $U$. More specifically, there exists $I \subsetneq \{1,...,p\}$ such that for $l \in \{1,...,\Dtau\}$ $\tilde{\mu}_{\tau_l}^{(i)} \neq 0$ for all $i  \in I$, and $\tilde{\mu}_{\tau_l}^{(i)} = 0$ otherwise.
    \item $S$ is a normal matrix with all its eigenvalues different, and $S$ remains constant throughout the observation and up to the time-horizon $T$. 
\end{enumerate}
\vspace{-0.5em}
The sparsity hypothesis is commonly used in the literature and can have interesting applications in our context since it provides insights about the nature of a transition. For example, if we suppose that $S$ is the Laplacian matrix of the graph $G$, then its first eigenvector (the one corresponding to the smallest eigenvalue) is a graph signal that is constant over the entire graph; then the next $k$ eigenvectors encode information about the cluster structure of the data, and $k$ depends on the complexity of this structure; the higher the eigenvalue, the higher the variance of the associated eigenvector \citep{Ortega2018}. The connection with our algorithm can be made as follows: suppose, we have computed the Graph Fourier coefficients of the means $\tilde{\mu}_{\tau_{l}}, \tilde{\mu}_{\tau_{l+1}} \in \real^p$, for the segments $\tau_{l}$ and $\tau_{l+1}$. Then, if $|\tilde{\mu}_{\tau_{l}}^{(i)}-\tilde{\mu}_{\tau_{l+1}}^{(i)}|$ is higher around the $i$-s (\ie dimensions in the spectral representation) corresponding to low frequencies, the difference implies a change occurring more uniformly across the graph, \eg a shift of the same magnitude at most of the nodes or affecting the nodes belonging to the same cluster. On the other hand, if $|\tilde{\mu}_{\tau_{l}}^{(i)}-\tilde{\mu}_{\tau_{l+1}}^{(i)}|$ is concentrated around $i$-s lying at high frequencies, the shift should show many sign changes across graph edges, indicating -for example- that a set of random nodes (\ie the formation of the set cannot be explained by the graph, \eg it is not a cluster of nodes) has modified their mean. The eigenvectors of the adjacency matrix also provide information about the connectivity of the graph as discussed in \cite{Ortega2018} and verified in a theoretical analysis in \cite{Athreya2017}.

As we suppose that $S$ does not change over time, the stationarity of the graph signals with respect to the graph implies that the covariance matrix $\Sigma_y=U \diag(P_y) U^*$ remains unchanged too. Then the average log-likelihood of the SGS can be written as:
\begin{equation}{\label{eq:loglikelihood}}
\begin{aligned}
   \!\!\!L(\mu,\tau) 
    & =-\!\sum_{l=1}^{\Dtau} \sum_{t=\tau_{l-1}+1}^{\tau_{l}} \!\!\!\! {\textstyle \frac{(y_t-\mu_{\tau_l})^T \Sigma^{-1} (y_t-\mu_{\tau_l})}{2 T }+\frac{\log det (\Sigma)}{2 T}} \\
    & = -\!\sum_{l=1}^{\Dtau} \sum_{t=\tau_{l-1}+1}^{\tau_{l}} \sum_{i=1}^{p}\! {\textstyle \ \frac{(\tilde{y}^{(i)}_t-\tilde{\mu}^{(i)}_{\tau_l})^2}{2 T P^{(i)}_y}+\frac{\log P^{(i)}_y}{2 T}}.
\end{aligned}
\end{equation}
This formulation can be seen as a way to \emph{translate the signal from the vertex domain to the spectral domain} where the sample becomes independent %\NOTE{with respect}\mN{remove?} 
to the graph structure (see \Property{Property:independence}). It is true that we could decorrelate the data via other techniques such as  PCA or SVD, however this way we would lose the interpretability of the results with respect to the graph.% structure.

Based on \Eq{eq:loglikelihood}, we can define the following $\ell_1$-penalized least squares cost function: %cost function , which %Above, $C_{T} (\tau,\tmu,\tilde{Y})$ 
%represents the 
\begin{equation}\label{eq:cost-func}
\begin{aligned}
C_{T} (\tau,\tmu,\tilde{Y}) &= \sum_{l=1}^{\Dtau} \sum_{t=\tau_{l-1}+1}^{\tau_{l}} \sum_{i=1}^{p}  {\frac{(\tilde{y}^{(i)}_t-\tmu^{(i)}_{\tau_l})^2}{T P_y^{(i)}}} \\
 & \ \ \ + \sum_{l=1}^{\Dtau} \lambda_{l} \frac{\sum_{i=1}^{p} I_l | \tmu^{(i)}_{\tau_l}|}{T}, 
\end{aligned}
\end{equation}
where $\lambda_{l}$ is the penalization constant leading to the desired sparsity of the GFT that a priori is segment-specific, $I_{l}=\tau_{l}-\tau_{l-1}$ is the length of the $l$-th segment.

The hypothesis that the data follow a multivariate Gaussian distribution is actually only required to derive our theoretical results. However, looking at \Eq{eq:cost-func} we can see that it just involves a least squared error (LSE) term standardized by the PSD, and therefore there is no requirement for Gaussian data. We also verify empirically in our experiments (\Sec{sec:exps}) that we can still get good performance in practice even when this data property does not hold.

%It is in fact not necessary that the data follows a multivariate Gaussian distribution since in \Eq{eq:cost-func} there is just a least squared error (LSE) term standardized by the PSD. The Gaussian hypothesis is actually only required for our theoretical results, but even if this is not the case, we can still get a good performance in practice (see the experiments in \Sec{sec:exps}). 

\subsection{Penalized cost function for an SGS with sparse GFT representation.}

\inlinetitle{Lasso-based GS change-point detector (LGS)}{.} %We can define a cost function \NOTE{inspired in} \Eq{eq:loglikelihood} and recover the change-points by minimizing it.%
Many graph signals observed in real applications can be accurately approximated by a subset of Graph Fourier frequencies, it is necessary to further exploit this feature when computing the means of each segment \citep{Perraudin2017,Marques2017,Huang2016}. This justifies adding an $\ell_1$-penalization term to \Eq{eq:loglikelihood}. Furthermore, to address the issue of the unknown number of change-points we also add the penalization term $pen(d)$.

The overall optimization problem for the change-point detection is written as: %
\begin{equation}{\label{eq:Change-point_detection_problem}}
\begin{aligned}
%\!\!\!\!\!\!\!\!\!(\hat{d}, \hat{\tau}(\hat{d}), \cpmu(\hat{d}))&= 
&(\hat{d}, \{\hat{\tau}_0,\hat{\tau}_1,...,\hat{\tau}_{\hat{d}} \},
\{\hat{\tmu}_1,...,\hat{\tmu}_{\hat{d}} \} ) \\
&:= \argminA_{\substack{d \in \{1,...,T\} \\ \tau \in \Tau^{d}  \\ \hat{\mu} \in F_{\tau}}} C_{T} (\tau,\tmu,\tilde{Y}) + pen(d)           \\
&= \argminA_{\substack{d \in \{1,...,T\} \\ \tau \in \Tau^{d}}} \sum_{l=1}^{d}  \argminA_{\hat{\mu} \in F_{\tau}}  \!\!\left[ \sum_{t=\tau_{l-1}+1}^{\tau_{l}} \sum_{i=1}^{p}  {\frac{(\tilde{y}^{(i)}_t-\tmu^{(i)}_{\tau_l})^2}{TP_y^{(i)}}} \right. \\
& \ \ \ \ + \left. \lambda_{l} \frac{\sum_{i=1}^{p} I_l | \tmu^{(i)}_{\tau_l}|}{T}  \right]   +\frac{d}{T}\left(c_1+c_2 \log\frac{T}{d}\right),\\
\end{aligned}%\!\!\!\!\!\!\!\!\!\!
\end{equation}

where $\Tau^{d}$ is the set of all possible segmentations of an SGS of size $d$.

\Problem{eq:Change-point_detection_problem} requires estimating the GFT of the mean of the graph signals that is assumed segment-wise constant. The separability of the cost function implies that, each time, this parameter depends only on the observations belonging to one segment, which is delimited by its change-points. This leads to the closed-form solution for ${\bar{\tmu}}_{\tau_l}^{(i)}$: 
\begin{equation}\label{eq:mean}
{\bar{\tmu}}_{\tau_{l}}^{(i)}  = \operatorname{sign}\left({\bar{\tilde{y}}}_{\tau_l}^{(i)}\right)\  * \ 
\max\left( {\left|{\bar{\tilde{y}}}_{\tau_l}^{(i)}\right|}- \frac{\lambda_l P_y^{(i)}}{2} , 0 \right)\!,
\end{equation}
where ${\bar{\tilde{y}}}_{\tau_l}^{(i)} = \frac{1}{I_l}\sum_{t=\tau_{l-1}+1}^{\tau_{l}}\tilde{y}_{t}^{(i)}$. %
%\begin{equation}{\label{eq:mean}}
%\begin{aligned}
%{\bar{\tmu}}_{\tau_{k}}^{(i)} & = \operatorname{sign}\left({\frac{\sum_{t=\tau_{l-1}+1}^{\tau_{l}}\tilde{y}_{t}^{(i)}}{I_l}}\right) * \\
%& \max\left( {\left|\frac{\sum_{t=\tau_{l-1}+1}^{\tau_{l}}\tilde{y}_{t}^{(i)}}{I_l}\right|}- \frac{\lambda_l P_y^{(i)}}{2} ,0 \right)\!\!.
%\end{aligned}
%\end{equation}
%\begin{equation}\label{eq:mean}
%{\bar{\tmu}}_{\tau_{k}}^{(i)} = \operatorname{sign}\!\!\left({\frac{\sum_{t=\tau_{l-1}+1}^{\tau_{l}}\tilde{y}_{t}^{(i)}}{I_l}}\right) * \max\!\!\left( {\left|\frac{\sum_{t=\tau_{l-1}+1}^{\tau_{l}}\tilde{y}_{t}^{(i)}}{I_l}\right|}- \frac{\lambda_l P_y^{(i)}}{2} ,0 \right)\!\!.
%\end{equation}
%
%
%Thanks to this formulation, it is easy to see how we can build and algorithm to find the precise change-points using dynamic programming (see \Algorithm{alg:l1changepointdetector}). %The final algorithm can be found in \Algorithm{alg:l1changepointdetector}.
Thanks to this formulation, we can make use of dynamic programming in \Algorithm{alg:l1changepointdetector} to find the precise change-points.
 
Even though \Problem{eq:Change-point_detection_problem} is computationally easy to solve, it still requires to set the $\lambda_{l}$ parameter, related with the sparsity of the graph signals, and a penalization term $pen(d)$ that would allow us to infer the number of change-points. This problem is not trivial since the number of possible solutions depends on the time-horizon $T$ and the number of nodes $p$; this feature hinders an asymptotic analysis. %We require penalization terms that have good performance in practice and depend on $p$ and $T$. 
Following the model selection approach, we can obtain an oracle-type inequality for the estimators $\hat{\tau}(\hat{d})$ and
$\cpmu(\hat{d})$. Nevertheless such analysis allows us to only infer the shape of $pen(d)$ depending on the unknown constants $c_1$, $c_2$ and a lower bound for $\lambda_{l}$. In model selection, it is common to infer the penalization parameters by the slope heuristics: an approach that is based on the idea that the minimum amount of penalization before overfitting can be observed, this is called \emph{minimal penalty} \cite{Baudry2010,Arlot2019_b}. Unfortunately, in our context it is not trivial to define the minimal penalty, hence we propose the VSGS change-point detector.  

\inlinetitle{Variable Selection-based GS change-point detector (VSGS)}{.} The idea of the detector (\Alg{alg:modelselectionchangepointdetector}) is to estimate the parameter $\lambda$ from a grid of candidates $\Lambda$, and the equivalents of the penalization constants $c_1$ and $c_2$ via the slope heuristic. To this end it defines the auxiliary optimization \Problem{Eq:optimization_problem_2} made of three terms: the first one aims to reduce the LSE, the second encourages sparse solutions based on the elements of $\Lambda$, the third one aims to determine the right number of change-points. We denote by $\LSEd$, $\LSEtau$, and $\LSEmu$ the solutions to this auxiliary problem:
\begin{equation}{\label{Eq:optimization_problem_2}}
\begin{aligned}
& (\LSEd,\LSEtau ,\LSEmu) \\
& :=\argmin_{\substack{ d \in {1,..,T} \\ \tau \in \Taud, \\ \tmu \in S_{(D_m,\tau)}}} C^{\text{LSE}}_T(\tau,\hat{\mu},\hat{Y}) + pen(m,\tau) \\
&=\argmin_{\substack{ d \in {1,..,T} \\ \tau \in \Taud, \\ \tmu \in S_{(D_m,\tau)}}}  \Bigg\{
\sum_{l=1}^{d} \Big( \sum_{t=\tau_{l-1}+1}^{\tau_{l}} \sum_{i=1}^{p}  {\frac{(\tilde{y}^{(i)}_t-\tmu^{(i)}_{\tau_l})^2}{TP_y^{(i)}}}\Big)
\\ &  \ \ \ \ \ \ \ \ \ \ \ \ \ \ \ \ +K_1\frac{D_{m}}{T} +\frac{d}{T}\left(K_2+K_3 \log\frac{T}{d}\right)\!\Bigg\}.
\end{aligned}
\end{equation}
\Problem{Eq:optimization_problem_2} should be minimized over sets of the form: 
\begin{equation}
\!\!\!\!\!\!\!S_{(D_m,\tau)}:=\{ \tmu\!\in\!F_{\tau}\ |\ \tmu_{\tau_l}\!\in\!S_{D_m}, %\text{ for all } 
\forall
l\!\in\!\{1,...,\Dtau\}\},\!\!\!
\end{equation}
where we denote by $S_{D_m}$ the space generated by $m$ specific elements of the standard basis of $\R^p$. In simple terms, we restrict the means defined in each of the segments to be elements of $S_{D_m}$. %
The quality of the solutions of \Problem{Eq:optimization_problem_2} is measured by comparing against an oracle inequality in \Theorem{Th:variable_selection_oracle}, and is further discussed in \Sec{sec:Model_selection}.

The full description of this new algorithm can be found in \Alg{alg:modelselectionchangepointdetector}. Note that for each element of $\Lambda$ a level of sparsity $D_{m_{\lambda}}$ is induced (Step 5). Then we use the auxiliary optimization \Problem{Eq:optimization_problem_2} to recover the right number of change-points and the level of sparsity $D_{m_{\lambda}}$ (Steps 6-11) . We use this information to recover the right parameter $\lambda$. The final solution will be the partition found via the auxiliary optimization problem and the means obtained after applying \Eq{eq:mean}.

\begin{algorithm}[t]
\small
\SetAlgoLined
\SetKwInOut{Input}{Input}
\SetKwInOut{Output}{Output}
\Input{$Y$: matrix $\R^{T \times p}$ representing the SGS \\
$w$: length of the warming period \\
$\Dtau_{\max}$: maximum number of change-points  \\
$U$: eigenvectors of the GSO \\
$(c_1,c_2)$: constants of the  penalization $pen(\tau)$ \\
$\lambda$: constant of the penalization $pen(\mu_{\tau})$ } 
\Output{$\hat{\tau}(\hat{d})$: set of change-points \\
$\cpmu(\hat{d})$: matrix $\R^{\hat{d} \times p}$ with rows being the\\\ \ \ \ \ mean of each segment}
Estimate the GFT of the dataset $\tilde{Y}=Y U$

Compute an estimation of $P_y$ using $w$ observations

\For{$d \in \{1,...,\Dtau_{\max}\}$}  
{ 
Apply dynamic programming to solve:\ \  %$\displaystyle\hat{\tau}(d),\cpmu(d) := \argmin_{ \tau \in \Tau_T^d}\ \argmin_{ \tmu \in F_{\tau}} C_T(\tau,\tmu,\tilde{Y})$
$\displaystyle\hat{\tau}(d),\cpmu(d) := \argmin_{\substack{ \tau \in \Tau^d,\\  \tmu \in F_{\tau}}}\ C_T(\tau,\tmu,\tilde{Y})$
%
%\vspace{-3mm}
%\begin{equation*}
%\begin{aligned}
%\hat{\tau}(d),\cpmu(d) & := \argmin_{ \tau \in \Tau_T^d}\ \argmin_{ \tmu \in F_{\tau}} C_T(\tau,\tmu,\tilde{Y})\vspace{-3mm}
%\end{aligned}
%\vspace{-3mm}
%\end{equation*}

}
Choose: %
\vspace{-3mm}
\begin{equation}
{\!\!\!\!\!\!\!\!\!\!\!\!\!\!\!\displaystyle\hat{d} :=\!\!\!\argmin_{d=\{\!1,...,\Dtau_{\max}\!\}}} \!\!C_T(\hat{\tau}(d),\cpmu(d),\tilde{Y}) + \frac{d}{T} \left(c_1\!+\!c_2 \log\frac{T}{d}\right)\!\!\!\!\!\!
\end{equation}

Return $\hat{\tau}(\hat{d})$ and
$\cpmu(\hat{d})$
 \caption{Lasso-based GS change-point detector (LGS)}\label{alg:l1changepointdetector} 

\end{algorithm}

\begin{algorithm}[h]
\small
\SetAlgoLined
\SetKwInOut{Input}{Input}
\SetKwInOut{Output}{Output}
\Input{$Y$: matrix $\R^{T \times p}$ representing the stream of \\\ \ \ \ \ the graph signal \\ 
$\Dtau_{\max}$: Maximum number of change-points \\
$w$: length of the warming period \\
$U$: eigenvectors of the GSO \\
$\Lambda$: values of the penalty term  $pen(\mu_{\tau})$ \\
}
\Output{$\hat{\tau}(\hat{d})$: set of change-points \\
$\cpmu(\hat{d})$: matrix $\R^{\hat{d} \times p}$ with rows being the \\\ \ \ \ \ mean of each segment}
Estimate the GFT of the dataset $\tilde{Y}=Y U$

Compute an estimation of $P_y$ using $w$ observations

\For{$\lambda \in \Lambda$}  
{ 
Solve the Lasso problem: \ $\tmu_{\text{Lasso}}:= {\displaystyle\argminA_{\tmu \in R^{T \times p}}}\sum_{t=1}^T \sum_{i=1}^p \frac{\norm{\tilde{y}^{(i)}_t-\tmu^{i}}^2}{TP_y^{(i)}} + \lambda \norm{\tmu}_1$ \\
Define $D_{m_{\lambda}}:= \norm{\tmu_{\text{Lasso}}}_0$

\For{$d \in \{1,...,\Dtau_{\max}\}$}  
{
Solve the change-point detection problem via dynamic programming $\hat{\tau}(d,D_{m_{\lambda}}),\cpmu^`(d,D_{m_{\lambda}}) :=$
\begin{equation*}
\begin{aligned}
& \argminA_{({\tmu,\tau}) \in  S_{(D_{m_{\lambda}},\tau(d))}}\!\! 
C^{\text{LSE}} (\tmu(d,D_{m_{\lambda}}),\tau(d,D_{m_{\lambda}})) 
\\ 
\end{aligned}
\end{equation*}
}
}
Find $K_1,K_2,K_3$ using the slope heuristic

Solve: $(\hat{\lambda},\hat{d}) :=$%
\begin{equation}{\label{eq:op_problem_2}}
\begin{aligned}
&\!\!\!\!\!\!\!\!\!\!\!\!\argmin_{\substack{\lambda \in \Lambda,\\d=\{1,...,\Dtau_{\max}\}}}\textstyle C^{\text{LSE}}(\hat{\tau}(d,D_{m_{\lambda}}),\cpmu^{\text{LSE}}(d,D_{m_{\lambda}}))\\ & \ \ \ \ \ \ \ \ +K_1\frac{D_{m_{\lambda}}}{T} +\frac{d}{T}\left(K_2+K_3\log\frac{T}{d}\right) 
\end{aligned}
\end{equation}

Keeping the segmentation $\hat{\tau}(\hat{d},\hat{D}_{m_{\hat{\lambda}}})$ and $\hat{\lambda}$ fixed, recover $\cpmu(\hat{d})$ via \Eq{eq:mean}

Return $\hat{\tau}(\hat{d},\hat{D}_{m_{\hat{\lambda}}})$ and $\cpmu(\hat{d})$

\caption{Variable Selection-based GS change-point detector (VSGS) }\label{alg:modelselectionchangepointdetector}
\end{algorithm}

\inlinetitle{How to estimate $P_y$}{.} 
Both algorithms require the knowledge of $P_y$ of the SGS (Step 2 in \Alg{alg:l1changepointdetector} and \Alg{alg:modelselectionchangepointdetector}). %\NOTE{
The empirical covariance estimator of the GFT, $\hat{Y}$, has the disadvantage that its variance scales with the norm of $P_y$, which is a problem in high dimensions when the distance between change-points is not long enough%}
. There are several algorithms aiming to solve this problem (\cite{Marques2017}). In particular, the estimator proposed by \cite{Perraudin2017} requires a smaller number of samples and its computation scales with the number of edges in the graph (sparse in most applications). The idea of the estimator is based on \Property{prop:GF_stationarity}: once the vertex domain of a stationary graph signal is known, it is possible to use different filters to generate synthetic observations corresponding to different regions of the graph, and then use them to reconstruct the PSD. This enables a good approximation of the PSD even when having a single graph signal as input. This is the method we use in our experiments.

\section{Model selection approach}\label{sec:Model_selection}

The change-point detection problem for the mean of an SGS can be written as a generalized linear Gaussian model%\NOTE{after preprocessing the data and the hypothesis of Gaussianity}.With regards to the preprocessing,
: we will detect the change-points over the GFT of the SGS, that is $\tilde{Y}$ instead of $Y$; we have standardized $\tilde{Y}$ so that the variance of all the GFT coefficients to be $\epsilon = 1$. %
Under these conditions, we define an isonormal process $(W(\tmu))_{\tmu \in \R^{T \times p}}:W(\tmu):=\frac{\Tr({{\eta}^\transpose \tilde{\mu}})}{T}$, where $\eta \in \R^{T \times p}$ is a matrix whose rows follow a centered multivariate Gaussian distribution with covariance matrix $\mathbb{I}_p$. %
The generalized Gaussian process related to the SGS can be written as: 
\begin{equation}{\label{eq:GGP}}
\tilde{Y}_{\epsilon}(\tmu)=\frac{\tr{(\tmu^{*})^\Top \tmu}}{T}+\epsilon W(\tmu). 
 \end{equation}
This formulation enables the use techniques from the model selection literature  \citep{Massart2003} in order to design the penalized term $pen(d)$ related with the number of change-points, and derive oracle-type inequalities for the performance of the proposed estimators described in \Alg{alg:l1changepointdetector} and \Alg{alg:modelselectionchangepointdetector}.

\Theorem{Th:main_result} is an \emph{oracle inequality} that provides insights on how \Alg{alg:l1changepointdetector} behaves with respect to the time-horizon $T$ and the graph size $p$. Furthermore, it gives us a guideline towards choosing $\lambda_l$ and the number of change-points in order to minimize the penalized mean-squared criteria, which is one of the differences of our work to the change-point detection algorithms analyzed in \cite{Lebarbier2003,Arlot2019} that are based in model selection too, but they focus on mean-squared criteria. 

\begin{theorem}{\label{Th:main_result}}
 Assume that:
\begin{equation}{\label{eq:constants}}
\begin{aligned}
\lambda_{l}&=\lambda \geq \frac{(3\sqrt{2}) \epsilon \sqrt{\log p + L}}{T}, \\ pen(\Dtau)&=\frac{\Dtau}{T}\left(c_1+c_2 \log\frac{T}{\Dtau}\right),
\end{aligned}
\end{equation}
where $c_1 \geq 6 \sqrt{2} \epsilon^2$, $c_2 \geq 3 \sqrt{2}\epsilon^2$, and $L$ is such that $L > \log 2 $. Then, there exists an absolute constant $C>0$ satisfying:
\begin{equation}{\label{inq:oracle_main_result}}
\begin{aligned}
\!\!\!\!\!\Exp\left[\frac{\norm{\cpmu-\tmu^*}^2_F}{T}+\lambda  \norm{\cpmu}_{[\hat{\tau}]}\!\!+pen(\hat{d})\right] 
\leq \,%
 \\ C(K) \left[\Bigg(\!\inf_{\tau \in \Tau} \bigg(\!\inf_{\substack{\tmu \in F_{\tau} \\ \norm{\tmu}_{[\tau]}<+\infty}}\!\!\!\! \frac{\norm{\tmu-\tmu^*}^2_F}{T} +\lambda \norm{\tmu}_{[\tau]}\bigg)\right. 
\\
\!\!\!\!\!\!\! +\,pen(d_\tau) \Bigg) +  2 \lambda \epsilon + \left(\!\!  1+\frac{1}{(e^{\gamma}-1)(e-1)}\right)\epsilon^2 \Bigg]\!, 
\end{aligned}
\end{equation}
where $\norm{\tmu}_{[\tau]}:=\frac{1}{T}\sum_{l=1}^{\Dtau} I_{\tau_l} \norm{\tmu_{\tau_l}}_1$, $\Tau$ is the set of all possible segmentations of the SGS, $\gamma= \frac{1}{K}(\sqrt{\log p+ L}-\sqrt{\log p + \log{2}})$, and $K>1$ is a given constant.
\end{theorem}

The proof follows similar arguments to \cite{Massart2011} and can be found in Appendix \ref{appendix_proof_Offline_detection}. Specifically, it requires first to define the set of models of interest, namely the list of candidate models indexed by the possible segmentations and $\ell_1$-balls of length $m \epsilon$, where $m \in \mathbb{N}^*$. 

The following lemma is a direct consequence of Corollary 4.3 in \cite{Giraud2015}.

\begin{lemma}{\label{lem:sparsity}} 
For any $L > 0$, the estimator obtained by solving \Problem{eq:Change-point_detection_problem} with tuning parameter
\begin{equation}
\!\!
    \lambda = 3 \epsilon \sqrt{2(\log p+L)},
\end{equation}
fulfills with probability at least $1-e^{-L}$ the risk bound:
\begin{equation}{\label{ineq:sparsity}}
\begin{aligned}
\!\!\frac{\norm{\cpmu - \tmu^*}_F^2}{T} 
& \leq \sum_{l=1}^{\Dtau} \sum_{t=\tau_{l-1}+1}^{\tau_l} \inf_{\substack{\tmu \neq 0,\\ \tmu \in R^p}} \frac{\norm{\tmu - \tmu_t^*}_2^2}{T} \\ & +\frac{18\epsilon^2(L+\log p)}{ T \Phi(\tmu)^2} \norm{\tmu}_0,
\end{aligned}
\end{equation}
where $\Phi(\tmu)$ is broadly known %in the literature 
as compatibility constant.
\end{lemma}

Both results, \Theorem{Th:main_result} and \Lemma{lem:sparsity}, provide details of the performance of the algorithm, when applied in practice, with respect to $\lambda$. The combination of both results suggests that \emph{the value of $\lambda_l$ should be the same for all the segments} in order to recover the sparsity of the signal. We can see that there is a trade-off between the performance of the estimator and its ability to recover the sparsity of the signal: On one side, we need a low $\lambda$ value in order to reduce the bias of the estimator (see \Ineq{inq:oracle_main_result}), while on the other side we need a high $\lambda$ value that will allow us to recover the sparsity of the signal with higher probability \Ineq{ineq:sparsity}. 

\Theorem{Th:main_result} provides lower bounds for the values of $c_1$ and $c_2$. Nevertheless, in practice, when fixing $c_1$ and $c_2$ at these values, the obtained results were not satisfying. Finding the right constants in model selection is a common difficult problem %, which is actually a difficult problem as studied in 
\citep{Arlot2019_b}. The slope heuristic recovers the constants
using a linear regression of the empirical risk against the elements of a penalization term. However, the curve defined by the cost function including the $\ell_1$ term does not tend to remain constant as the number of change-points increases, a feature that is used in the slope heuristics. 

In \Algorithm{alg:modelselectionchangepointdetector}, we replace the $\ell_1$-penalization term by a Variable Selection penalization term. For each of the elements of a given set of penalization parameters $\Lambda$, we solve a  Lasso problem over the whole SGS. This allows us to keep all the relevant frequencies. Then, we solve multiple change-point detection problems for different levels of sparsity. We can deduce the right level of sparsity controlled by $K_1$, as well as the two constants $K_2$ and $K_3$ related with the number of change-points, via the slope heuristic. This approach is validated by \Theorem{Th:variable_selection_oracle} and the experiments in \Sec{sec:exps}.

\begin{theorem}{\label{Th:variable_selection_oracle}}

Let $\hat{\tau}$ and  $\cpmu$ be solutions to the optimization \Problem{Eq:optimization_problem_2}. Then, there exist constants $K_1$, $K_2$, $K_3$ defining the penalty term  for all $(m,\tau) \in M$, where $M \subset \{1,...,p\} \times \Tau$:%
\begin{equation}{\label{eq:constants_model_selection}}
\begin{aligned}
\!\!\!pen(m,\tau)=K_1 \frac{D_m}{T}+\frac{d_{\tau}}{T}\left(K_2+K_3 \log\frac{T}{d_{\tau}}\right)\!,
\end{aligned}
\end{equation}
there exists a positive constant $C(K)$, and $K>1$ a given constant, such that:
\begin{equation}
\begin{aligned}
 &  \Exp\left[\frac{\norm{\cpmu-\tmu^*}^2_F}{T}\right] \leq  C(K) \left[\inf_{(m,\tau) \in M } \inf_{\tmu \in S(D_m,\tau)} \frac{\norm{\tmu-\tmu^*}^2_F}{T} \right.  \\ &  + pen(m,\tau)  +   \left(1+\left(\frac{1}{(e^\gamma-1)(e-1)}\right)\right) \epsilon^2 \Bigg] \!,
\end{aligned}
\end{equation}
where $\Tau$ is the set of all possible segmentations of the SGS, $\gamma= \frac{1}{K}(\sqrt{\log p+ L}-\sqrt{\log p + \log{2}})$.
\end{theorem}

The proof of \Theorem{Th:variable_selection_oracle} is a consequence of Theorem 4.18 in \cite{Massart2003}, and is provided in Appendix \ref{appendix_proof_Offline_detection}.

\begin{figure*}[t]\centering
\begin{minipage}[t]{0.03\linewidth}
	\rotatebox{90}{\colorbox{gray!15}{\qquad\qquad\qquad\qquad\quad\ \ GROUND TRUTH\qquad\qquad\qquad\qquad\quad\ \ }}
\end{minipage}
\begin{minipage}[h]{.90\linewidth}
  \centering
	\vspace{-28em}
  \ \ \ \ \ \ \ \ \ \ \ \ \ \includegraphics[width=0.20\linewidth, viewport=170px 45px 1090px 1700px, clip]{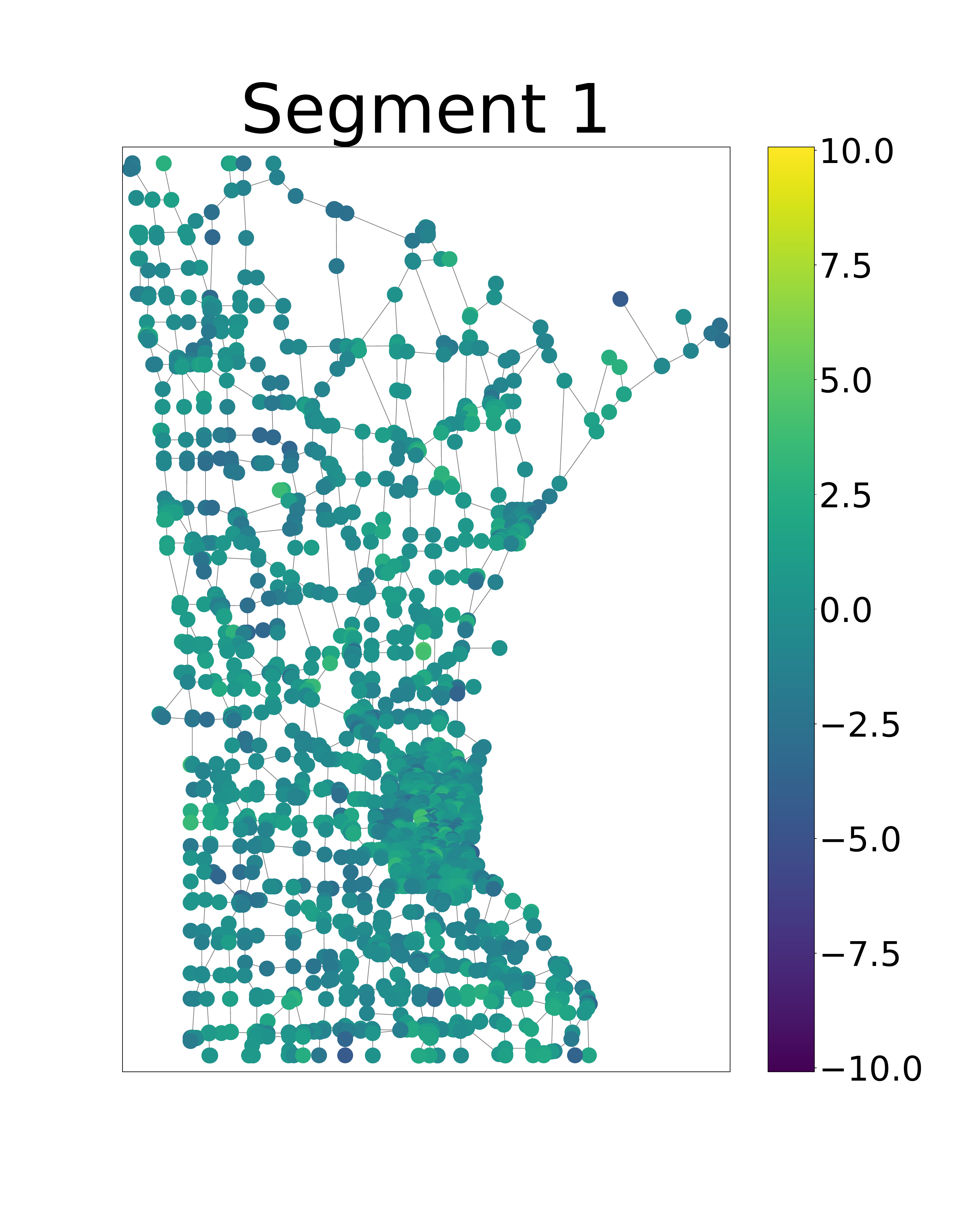}%
  \hspace{2em}
   \includegraphics[width=0.20\linewidth, viewport=170px 45px 1090px 1700px, clip]{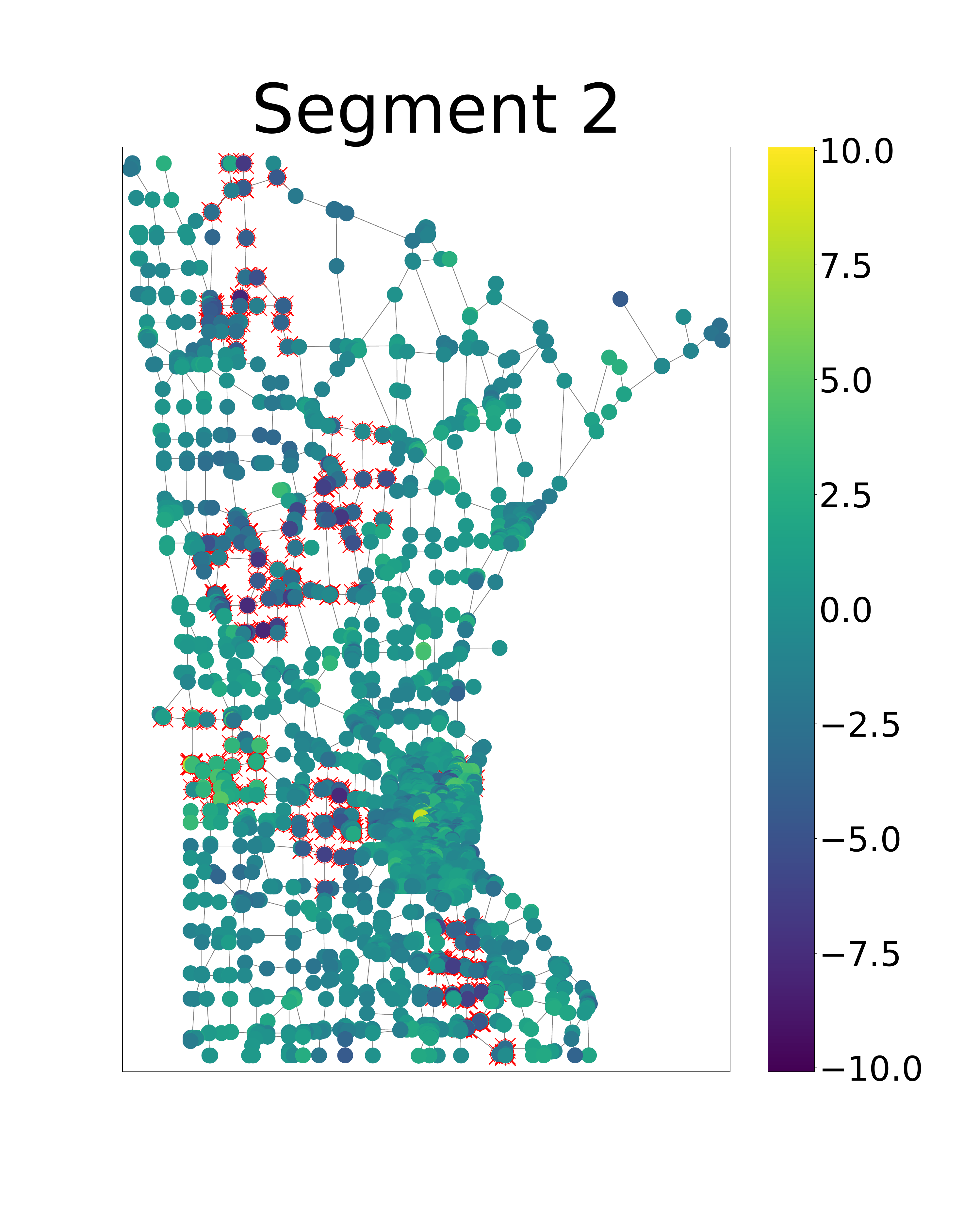}%
   \hspace{2em}
    \includegraphics[width=0.257\linewidth, viewport=170px 45px 1350px 1700px, clip]{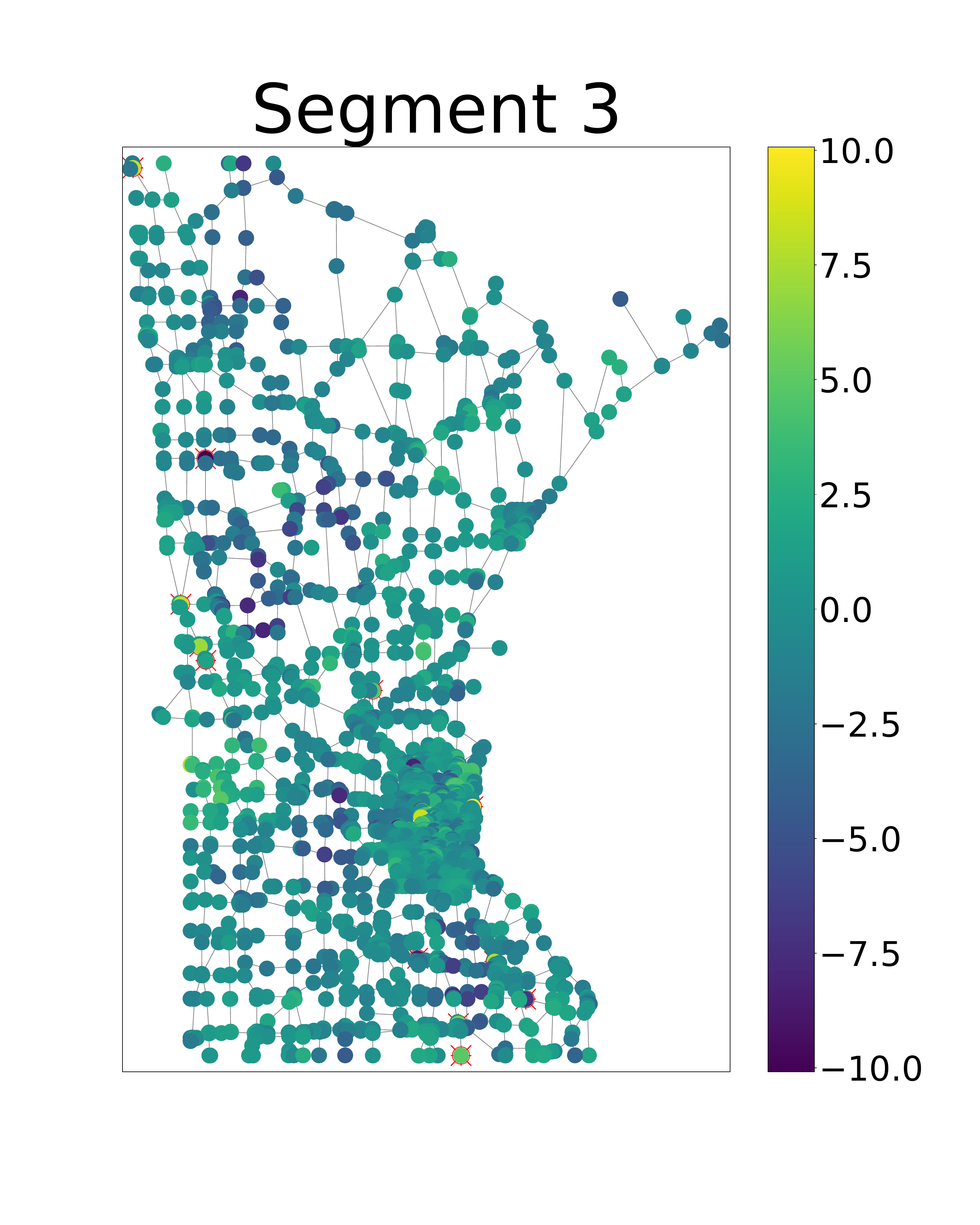} 
    \hspace{2em} \\
    \includegraphics[width=0.22\linewidth]{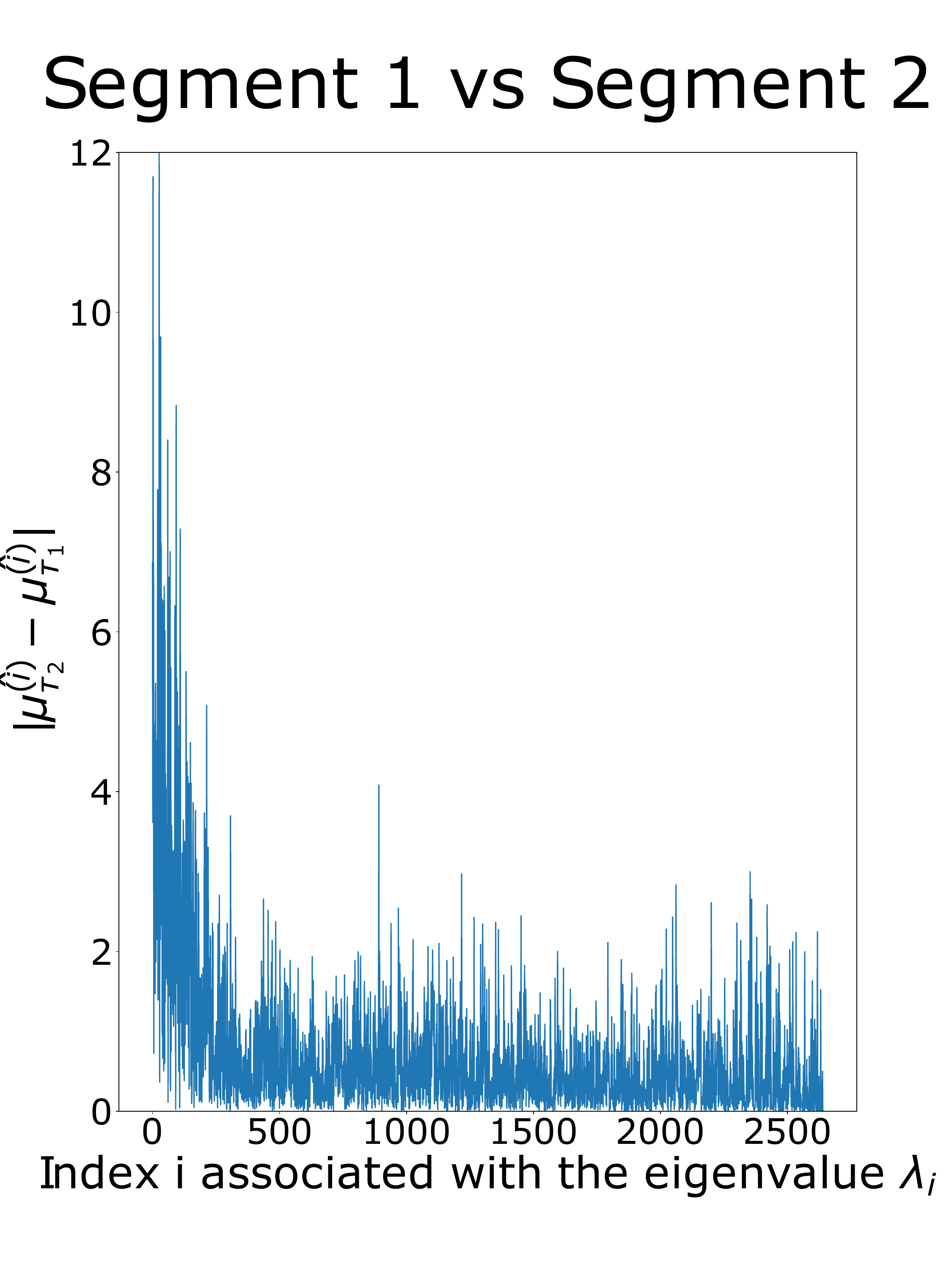}%
 %%\hspace{2em}
   \includegraphics[width=0.22\linewidth]{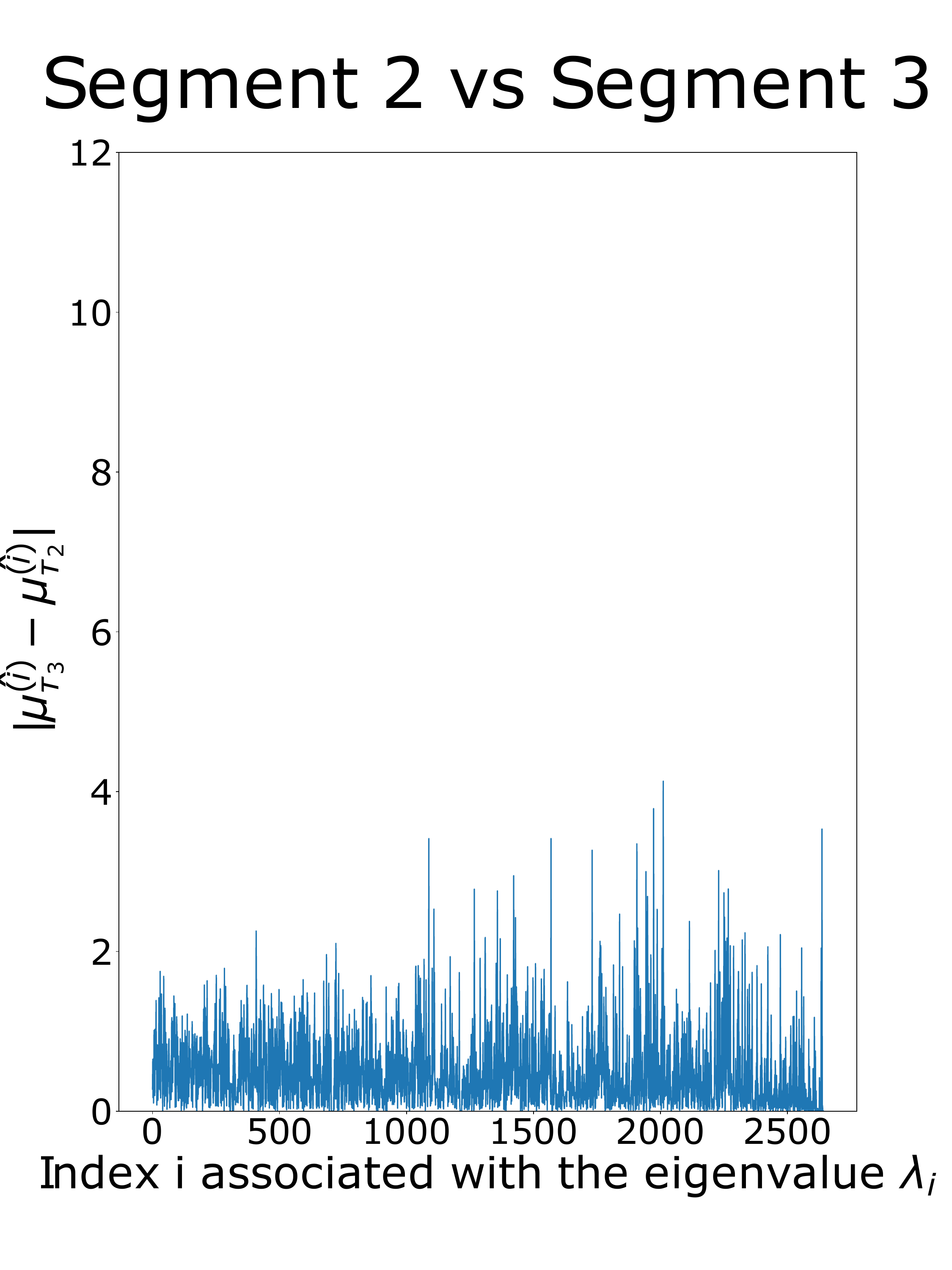} \\
\end{minipage} 
\vspace{1em}%
 \hrule
\vspace{1.3em}%
\begin{minipage}[t]{0.03\linewidth}
	\rotatebox{90}{\colorbox{gray!15}{\qquad\qquad\qquad\qquad\quad DETECTION RESULT\qquad\qquad\qquad\qquad\quad}}
\end{minipage}
\begin{minipage}[t]{.90\linewidth}
  \centering
	\vspace{-29em}
  \ \ \ \ \ \ \ \ \ \ \ \ \ \includegraphics[width=0.20\linewidth, viewport=170px 45px 1090px 1700px, clip]{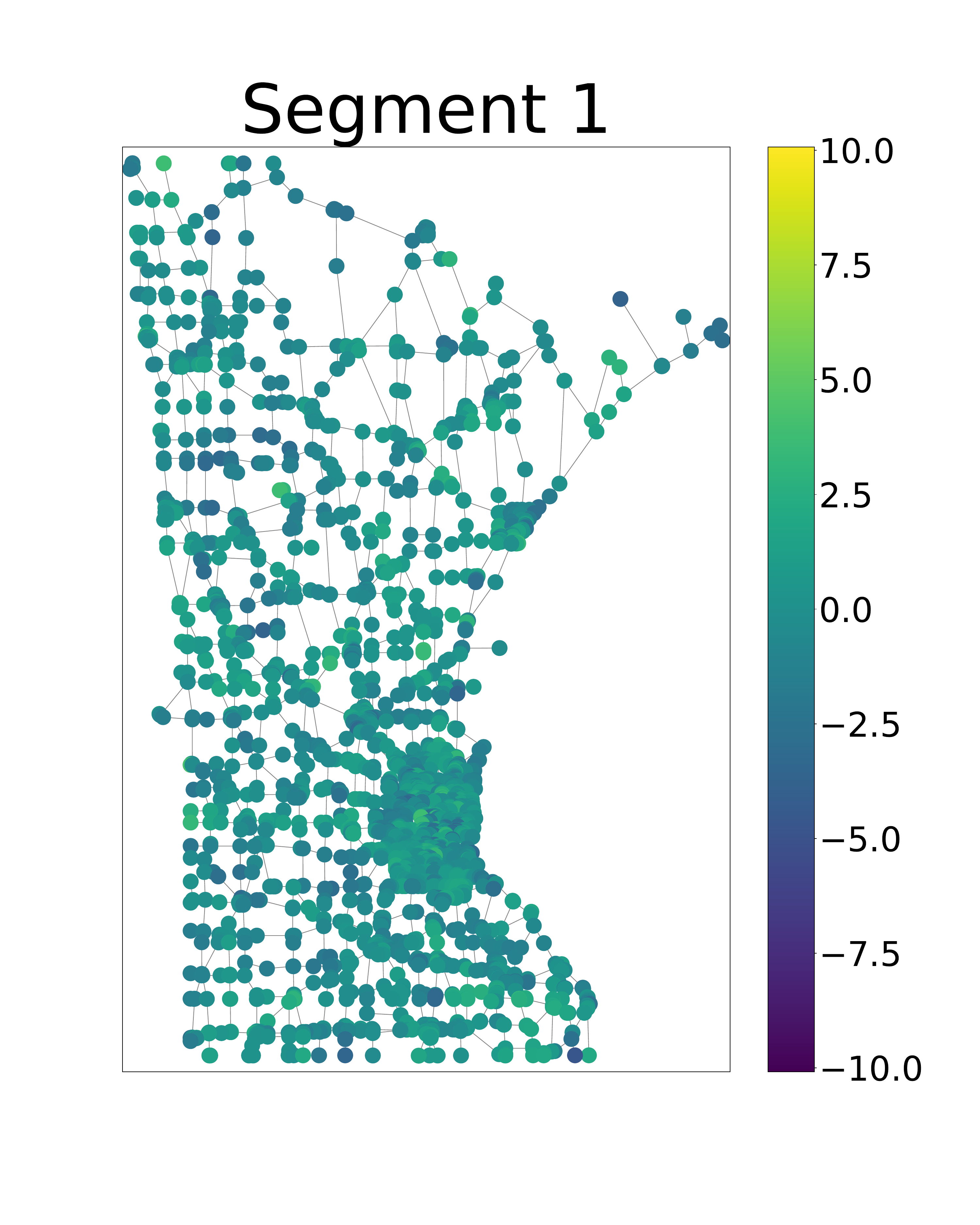}%
  \hspace{2em}
   \includegraphics[width=0.20\linewidth, viewport=170px 45px 1090px 1700px, clip]{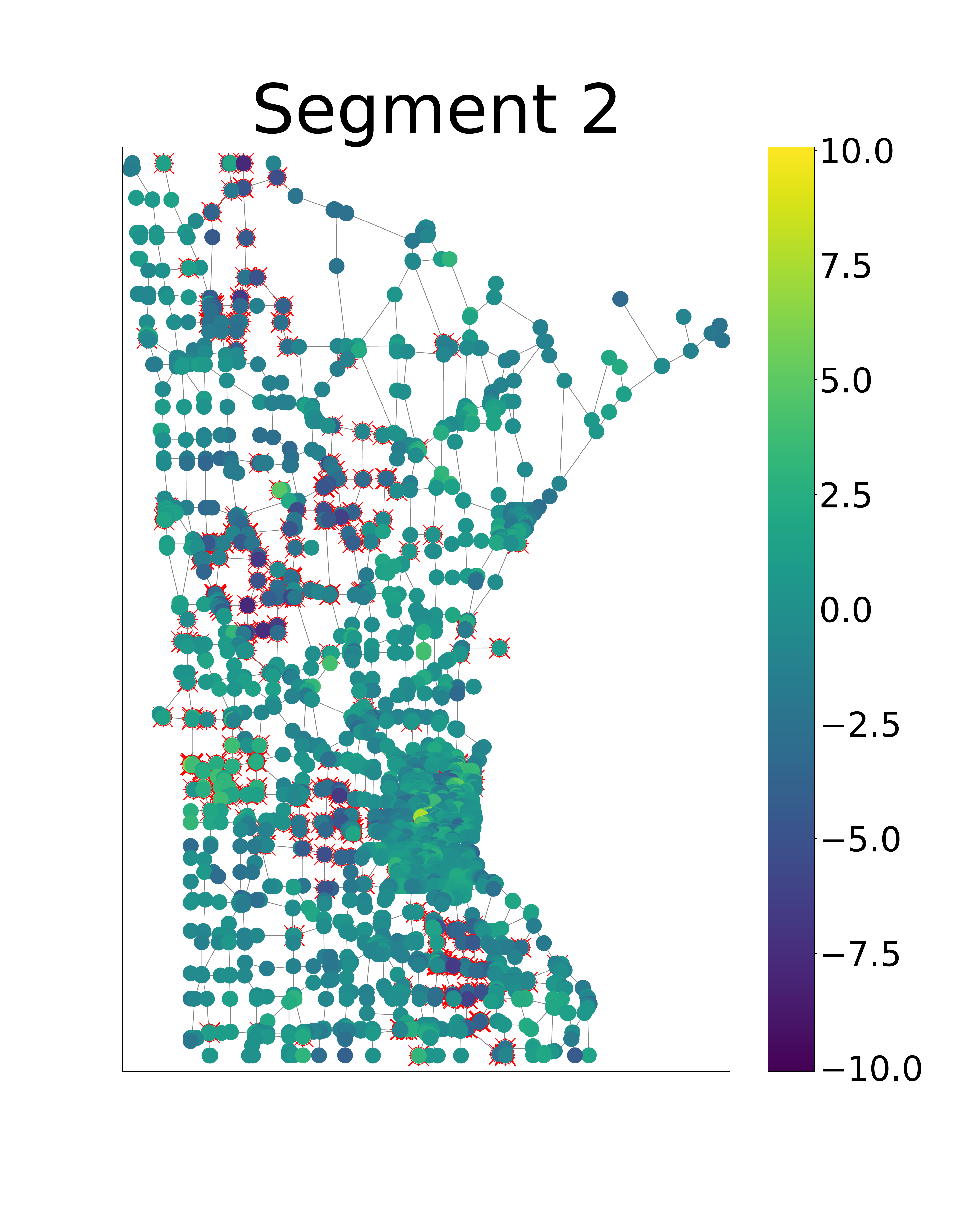}%
  \hspace{2em}
    \includegraphics[width=0.257\linewidth, viewport=170px 45px 1350px 1700px, clip]{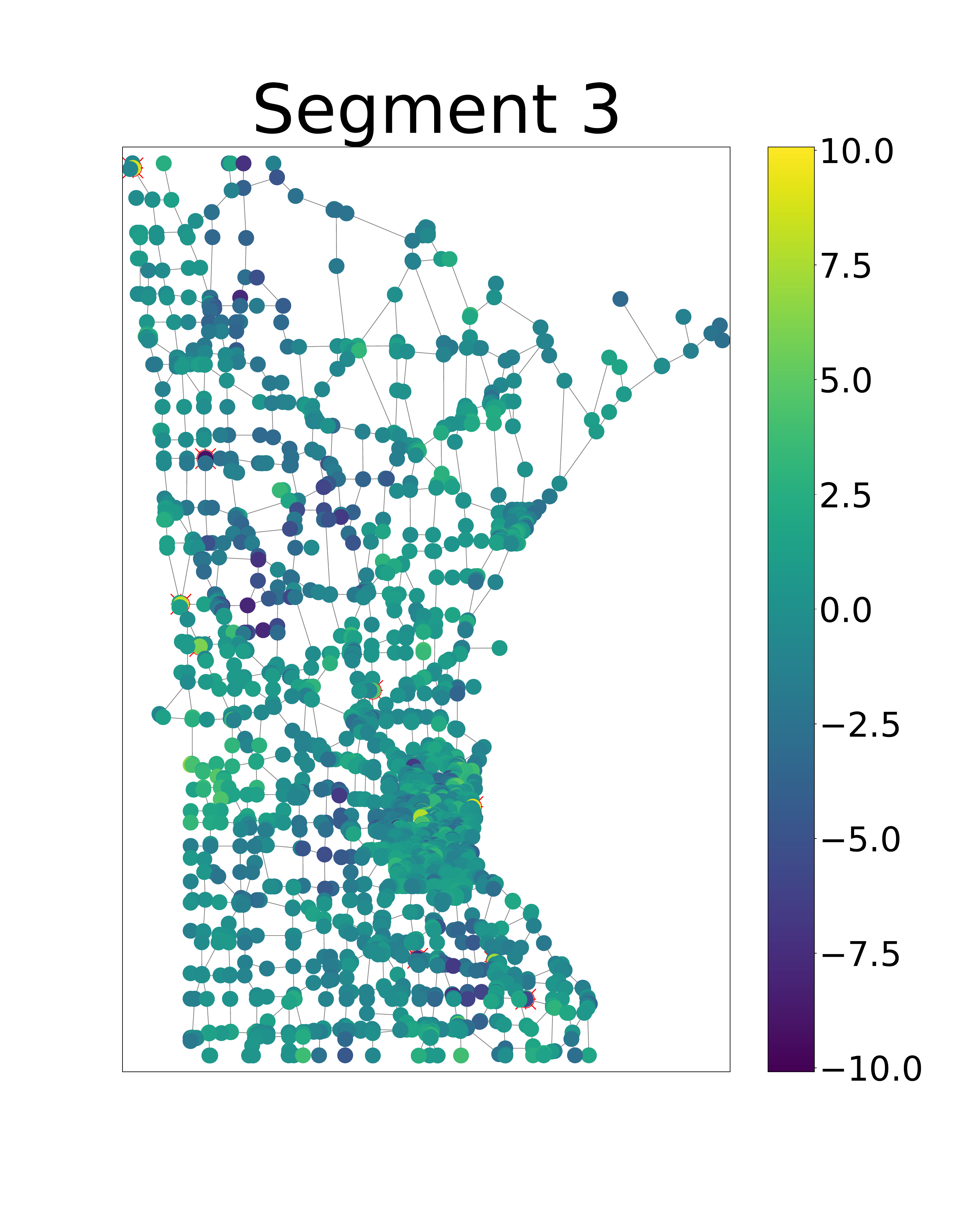}
 \hspace{2em} \\
    %\vspace{+1.0em}
    \includegraphics[width=0.22\linewidth]{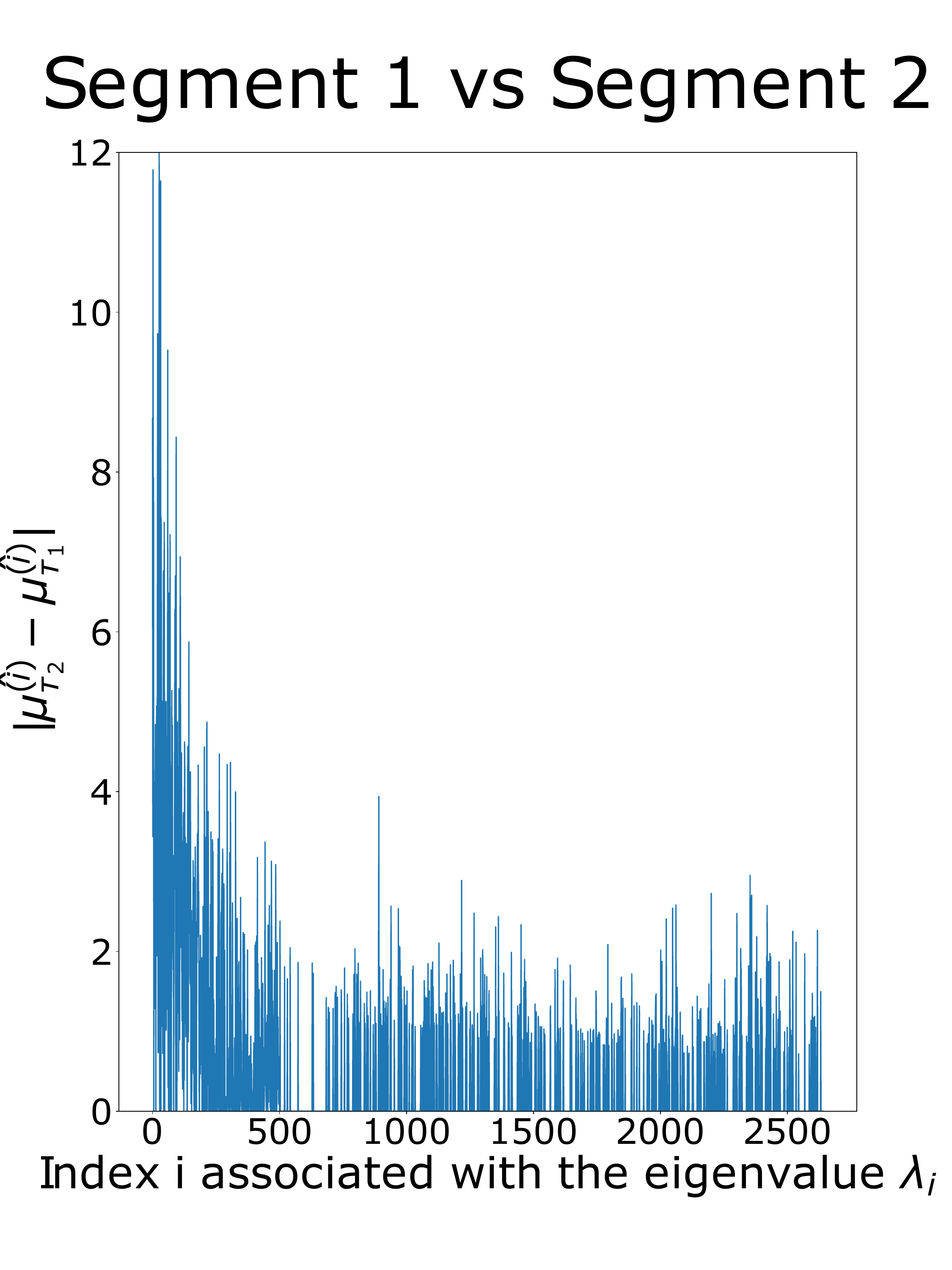}%
 %\hspace{2em}
   \includegraphics[width=0.22\linewidth]{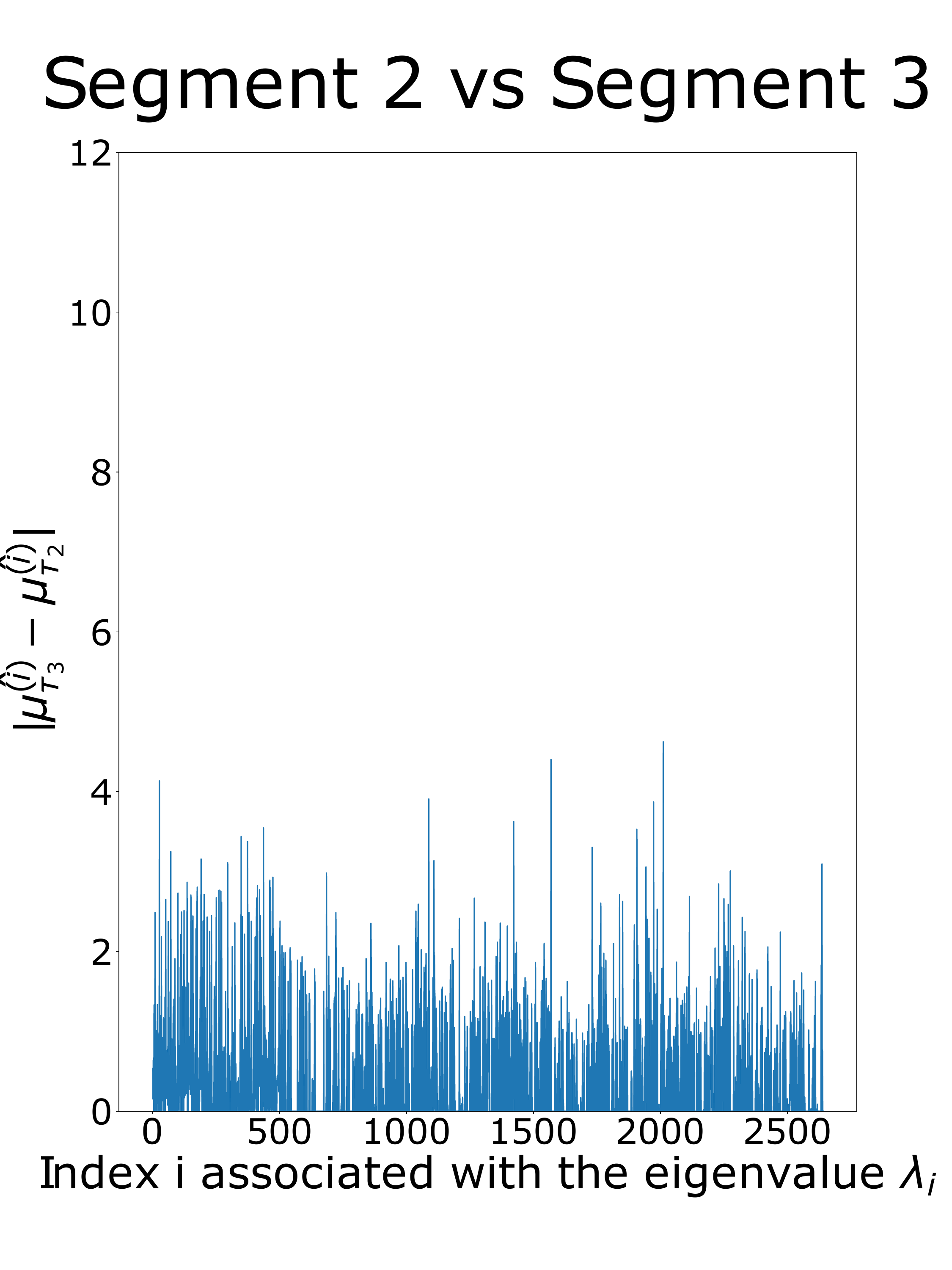}
\end{minipage}
\vspace{-0.5em}
\caption{ \small{Instance of Scenario III (after the first change-point: $10$ regions; after the second change-point: $20$ random nodes). The node colors indicate their expected values. \textbf{First two rows:} In the first row, the figures show the true mean in each of the 3 segments induced by the $2$ change-points. The red contour around a node indicates that its mean has changed compared to the previous segment. In the second row, the two plots show the true difference between the GFT of the mean of two consecutive segments (1st and 2nd, 2nd and 3rd).
\textbf{Last two rows:} The estimated mean is shown. The red contour around a node indicates that it was detected by our algorithm as having a change in its mean. In the last row, we show the difference between the GFT of the estimated mean of two consecutive segments.}
}
\label{fig:CP_Manhatan}
\end{figure*}

\section{Numerical experiments}{\label{sec:exps}}
\inlinetitle{Setup}{.} As mentioned earlier, a key hypothesis in our approach is the stationarity of the graph signals. An alternative definition states that a stationary graph signal is what we get as output after applying a graph filter $H$, with a frequency response $h(\theta)$, to white noise. This is equivalent to \Definition{Stationarity2} when the GSO is normal and all its eigenvalues are different \citep{Perraudin2017}. In this case we will use the Laplacian of the graph as GSO. We generate our graph signals via this definition. We design three different scenarios to test the capabilities of our method. For each scenario, we generate $100$ instances. 

\emph{Scenario I:} We generate Erd{\H o}s–Rényi (ER) graphs with a fixed link creation probability $p_{_{\operatorname{ER}}}=0.3$. The spectral profile of the filter is defined by: $h(\theta) \propto \frac{1}{\log{(\theta+10)}+1}$. The white noise generating the signal is a zero mean uniform distribution with variance 1.  We generate  change-points via a Poisson distribution with mean value $5$ and the distance between them is generated at random via $30+\epsilon$, where $\epsilon$ follows an exponential distribution with mean value equal to $20$ (the expected distance is 50). Before the first change-point, the mean of the graph signals is a linear combination of the first $20$ eigenvectors of the Laplacian matrix; $20$ random coefficients of the GFT are changed right after each change-point, the mean is then this new linear combination of eigenvectors. In all cases the coefficients of the linear combinations were generated uniformly at random in the interval $[-5,5]$.

\emph{Scenario II:} The graph structure is generated by a Barabasi-Albert (BA) model in which each incoming node is connected to $4$ nodes. The spectral profile of the filter is proportional to the density function of a Gamma distribution, $h(\theta) \propto p_{_{\Gamma(20,5)}}(\theta)$. The white noise generating the signal is a standardized Gaussian distribution. Then, $4$ change-points are generated and the distance between them is generated at random via $30+\epsilon$, with $\epsilon$ following an exponential distribution with mean value equal to $20$ (the expected distance is 50). Before the first change-point, the mean of the graph signals is a linear combination of the first $20$ eigenvectors of the Laplacian matrix; after the first change, the node with the highest degree and all its neighbors change their mean; after the second change-point the first $5$ nodes with the highest degrees modify their mean; after the third change-point, $20$ nodes at random of the graph get their mean changed. In all cases, the mean is generated uniformly at random in the interval 
$[-5,5]$.  

\begin{table*}[t]
\caption{Performance evaluation of the VSGS and Approximate VSGS change-point detectors for the synthetic Scenarios I, II, and III. The mean and (std) of the different evaluation metrics are estimated over $100$ generated instances of each scenario ($\uparrow$/$\downarrow$ indicate our preference for a higher/lower value for a metric). The Recall and the Precision are evaluated allowing a difference of $10$ timestamps between the estimated and the real change-points.}{\label{tab:synthetic experiments}} 
\footnotesize
%\scriptsize
\centering
\makebox[\linewidth][c]{%
%\begin{tabular}{c{0.5cm} l{0.5m} | r{0.5cm}  l{0.5cm} l{0.5cm} l{0.5cm} l{0.5cm}} %| C{2cm} | C{2cm} | @{}lc@{}
\scalebox{.883}{
\begin{tabular}{c r r | r l l l l }
    \toprule
    \textbf{Scenario} & \textbf{$\#$ Nodes} & \textbf{\ Detector} & \textbf{Hausdorff} ($\downarrow$) & \textbf{Rand} ($\uparrow$) & \textbf{Recall} ($\uparrow$) & \textbf{Precision} ($\uparrow$) & \textbf{\ \ F1} ($\uparrow$) \\
    \hline\hline
    \multirow{2}{*}{I} & \multirow{2}{*}{$100$} & 
    \ VSGS & $0.94 (00.24)$ \ \ & $0.99 (0.00)$ \ \ &  $1.00 (0.00)$  \ \ & $0.80 (0.15)$ \ \ & $0.88 (0.10)$ \\
    & & \  Approx. VSGS & $1.73 (70.88)$ \ \ & $0.99 (0.02)$  \ \ & $1.00 (0.05)$ \ \ & $0.88 (0.13)$ \ \ & $0.93 (0.08)$ \\
    \hline
    \multirow{2}{*}{II} & \multirow{2}{*}{$100$} & \ VSGS  & $0.84 (00.37)$ \ \  & $0.99 (0.01)$ \ \ & $1.00 (0.00)$  \ \ & $0.98 (0.07)$  \ \  &  $0.99 (0.04)$  \\
    & & \  Approx. VSGS  & $1.57 (07.28)$  \ \  &  $0.99 (0.01)$ \ \ & $1.00 (0.00)$ \ \ & $0.98 (0.07)$ \ \  & $0.99 (0.04)$
    \\
    \hline\hline
    \multirow{2}{*}{I} & \multirow{2}{*}{$500$} &
    \ VSGS & $0.94 (00.24)$ \ \ & $0.99 (0.00)$ \ \ & $1.00 (0.00)$  \ \ & $1.00 (0.00)$ \ \ & $1.00 (0.00)$ \\
    & & \  Approx. VSGS & $6.29 (17.13)$ \ \ & $0.98 (0.04)$ \ \ & $0.97 (0.11)$ \ \ & $1.00  (0.05)$ \ \ & $0.98 (0.09)$ \\
    \hline
    \multirow{2}{*}{II} & \multirow{2}{*}{$500$}
    & \ VSGS  & $10.43 (15.61)$ \ \  & $0.96 (0.05)$ \ \ & $0.91 (0.15)$  \ \ & $1.00 (0.00)$ \ \  & $0.94 (0.09)$ \\
    & & \  Approx. VSGS  & $12.48 (16.54)$ \ \  & $0.96 (0.06)$ \ \ & $0.89 (0.16)$ \ \ & $1.00 (0.00)$ \ \  &  $0.93 (0.09)$
    \\
    \hline
    \multirow{2}{*}{I} & \multirow{2}{*}{$1000$} &
    \ VSGS & $0.94 (00.23)$ \ \ &  $0.99 (0.00)$ \ \ & $1.00 (0.00)$  \ \ &  $1.00 (0.00)$ \ \ & $1.00 (0.00)$ \\
%    \hline
    & & \  Approx. VSGS & 
    $7.36 (23.46)$ \ \ & $0.98 (0.07)$ \ \ & $0.96 (0.12)$ \ \ & $1.00 (0.00)$ \ \ & $0.98 (0.09)$ \\
%    \hline
    \hline
    \multirow{2}{*}{II} & \multirow{2}{*}{$1000$} &
    \ VSGS & $33.02 (17.70)$ \ \  & $0.89 (0.06)$ \ \ & $0.71 (0.14)$ \ \ & $1.00 (0.00)$  \ \  & $0.83 (0.09)$   \\
%    \hline
    & & \  Approx. VSGS  &  $33.81 (17.05)$ \ \  & $0.89 (0.05)$  \ \ & $0.71 (0.12)$ \ \ & $1.00 (0.00)$  \ \ &  $0.83 (0.08)$
    \\
    \hline\hline
    III - $5$ rand. regions  & \multirow{2}{*}{$2642$} &
    \ VSGS & $119.48 (64.49)$ \ \ &  $0.82 (0.09)$ \ \ & $0.60 (0.02)$  \ \ &  $1.00 (0.00)$ \ \ & $0.73 (0.13)$ \\
%    \hline
     \ \ \ $10$ rand. nodes & & \  Approx. VSGS & $120.81 (63.38)$ \ \ & $0.82 (0.09)$ \ \ & $0.60 (0.20)$ \ \ & $1.00 (0.00)$ \ \ & $0.73 (0.13)$ \\
         \hline
    III - $10$ rand. regions  & \multirow{2}{*}{$2642$}
 & \ VSGS & $7.54 (29.82)$ \ \ &  $0.99 (0.04)$ \ \ & $0.98 (0.11)$  \ \ &  $1.00 (0.00)$ \ \ & $0.98 (0.07)$ \\
%    \hline
     \ \ \ \ \ $20$ rand. nodes & & \  Approx. VSGS & $8.85 (32.28)$ \ \ & $0.99 (0.04)$ \ \ & $0.97 (0.12)$ \ \ & $1.00 (0.00)$ \ \ & $0.98 (0.08)$ \\
      \hline
         III - $20$ rand. regions & \multirow{2}{*}{$2642$}
   & \ VSGS & $0.72 (00.45)$ \ \ &  $1.00 (0.00)$ \ \ & $1.00 (0.00)$  \ \ &  $1.00 (0.00)$ \ \ & $1.00 (0.00)$ \\
%    \hline
     \ \ \ \ \ $40$ rand. nodes & & \  Approx. VSGS  & $0.72 (00.45)$ \ \ &  $1.00 (0.00)$ \ \ & $1.00 (0.00)$  \ \ &  $1.00 (0.00)$ \ \ & $1.00 (0.00)$ \\
    \bottomrule
    %\hline\hline
\end{tabular}
}}
\end{table*}

\emph{Scenario III:} We generate a SGS over the Minnesota Road Network \citep{Rossi2015} that contains $2.6$k nodes, $3.3$k edges, and maximum $k$-core of 3 (\ie the spatial network is quasi-planar, see \Fig{fig:CP_Manhatan}). The spectral profile of the filter is defined by: $h(\theta) \propto \frac{1}{\log{(\theta+10)}+1}$. The white noise generating the signal is a Student-t distribution with $100$ degrees of freedom. We generate $3$ change-points and the distance between them is generated at random via $120+\epsilon$, with $\epsilon$ following an exponential distribution with mean value equal to $30$ (the expected distance is 150). In the first segment the mean of the graph signals is a linear combination of the first $500$ eigenvectors of the Laplacian matrix. %
Then, the first change is generated in a way that can be explained by the graph structure: %at the first change-point, 
a number of random non-overlapping regions %(5, 10, 20, or 40) 
are created, each one containing a random node and all nodes in its $5$-hop neighborhood that were not selected earlier. In each region, the direction of the signal change is kept same, either positive or negative for all of its nodes, and the intensity of the change is generated for each node uniformly at random in the interval [1,5]. %After the second change-point, 
Contrary% to the first change
, the second change does not respect the graph structure: random nodes are chosen and their value of their means is increased uniformly by a value in $[-10,-5] \cup [5,10]$. 

We proceed to the analysis of the performance of our method. Recall that our approach requires the PSD $P_y$ of the signal as a parameter. \Alg{alg:modelselectionchangepointdetector} is mentioned as VSGS when we use the real value of the $P_y$ and Approx. VSGS when we approximate it. For the later case, we estimate the $P_y$ of the signal with the technique of \cite{Perraudin2017} on the first $50$ graph signals of the SGS. 

We implement the slope-heuristic to recover the parameters $K_1$, $K_2$, and $K_3$: that is we make a linear regression of the cost-functions of the list of models with high complexity against the penalization terms, then we multiply the linear regression's coefficients by $-2$ as suggested in \cite{Arlot2019}. 

A code repository including the implementation of the proposed algorithms and the tested experimental scenarios is publicly available online.\footnote{\url{https://github.com/AlejandrodelaConcha/VSGS}}

\inlinetitle{Results}{.} Table 1 summarizes the results of our experiments by reporting several evaluation metrics \citep{truong2020}. The first thing to realize is that, generally, our method has good performance regardless of the PSD estimation or whether the Gaussian hypothesis holds, that is a the distance with respect to the real change-points (Hausdorff distance) is small given the minimum gap between change-points. Almost all the timestamps get correctly classified as whether they are change-points or not (Rand Index $\rightarrow 1$), the majority of the change-points are recovered (Recall $\rightarrow 1$) and not too many spurious change-points are wrongly identified (Precision $\rightarrow 1$). The effect of the PSD estimation can be seen in the Hausdorff distance, which tends to increase in mean and variance. 

%\NOTE{There is also an effect, when we increase the size of the graph while keeping the same average distance between change-points, which is clearer in the Scenario II where the Recall of the algorithm increases with the decrease of the number of nodes.}

\Fig{fig:CP_Manhatan} shows one instance of Scenario III, namely when the number of affected regions is $10$ and the number of random nodes modifying their mean is $20$. After the first change-point, as expected, it can be seen how the shift in the mean related to the structure of the graph makes $|\hat{\mu}_{\tau_{l}}^{(i)}-\hat{\mu}_{\tau_{l+1}}^{(i)}|$ higher in low frequencies. In this particular example, our algorithm was able to recover the right number of change-points and a good approximation of the means for each segment in both domains.
 
 %It also does a good job according to Table 1. 

The detailed evaluation metrics that are reported in \Tab{tab:synthetic experiments} confirm the quality of the obtained solutions. However, the intensity of the change-point is important for our algorithms to perform well: %
when a change is small in terms of the spectral domain, it is essentially harder to recover the respective change-points and the sparse representation. This can be seen in Scenario III where the performance of VSGS improves with the increase of the number nodes that get affected by the change. 

\section{Conclusion}% and future work}

%In this work 
We presented an offline change-point detection method for shifts in the mean of a stream of graph signals. The proposed approach infers automatically the number of change-points and the level of sparsity of the signal in its Graph Fourier representation. The formulation has the advantage of being easy to resolve via dynamic programming while also allowing interesting theoretical guarantees such as the oracle-type inequality that we provide.
The techniques and results of this paper could be generalized to situations where we aim to spot change-points in a stream of multivariate signals that supports a sparse representation in a orthonormal basis. 
%An important theoretical result that remains to be proven is the consistency of the change-points. 
Proving the consistency of the detected change-points and its generalization to types of graph signals representation other than the GFT, such as wavelets, is among our plans for future work.

\section*{Acknowledgment}
 This work was funded by the IdAML Chair hosted at ENS Paris-Saclay, Université Paris-Saclay and %the scholarship provided by 
 the DIM Math Innov network.

%\newpage

%%%% Add Scenario 33
%%%% Solve problem with table 
\clearpage
%\newpage
\balance
{\small
\bibliography{references}
}

\appendix

\onecolumn

\section{Proof of Theorems 1 and 2}{\label{appendix_proof_Offline_detection}}

We present the proofs of Theorem 1 and Theorem 2 of the main text. For completeness, we introduce key components such as basic concepts and results from the model selection literature.

The model selection framework offers an answer to the question: how to choose the function $pen(d)$ and the level of sparsity of the graph signals with respect to the Graph Fourier Transform (GFT) in order to guarantee good performance in practice of the proposed algorithms.

\begin{customdef}{5}
Given a separable Hilbert space $\Hil$, a generalized linear Gaussian model is defined by:
\begin{equation}{\label{eq:GLGM}}
Y_{\epsilon}(g)= \dott{f}{g}_{\Hil}+ \epsilon W(g), \  \text{  for all } \\ g \in \Hil, 
\end{equation}
where $W$ is an isonormal process (Definition \ref{def:Isonormal}).
\end{customdef}

\begin{customdef}{6}{\label{def:Isonormal}}
A Gaussian process $(W(g))_{g \in \Hil}$ is said to be isonormal if it is centered with covariance given by $\Exp[W(h) W(g)]= \dott{h}{g}_{\Hil}$ for all $h,g \in \Hil $.
\end{customdef}

An isonormal process is the natural extension of the notion of standard normal random vector to the infinite dimensional case. 

In the main text we said the change-point detection problem can be restated as a generalized linear Gaussian model, where $\Hil= \R^{T \times p}$: the dot product $\dott{h}{g}_{\Hil}$ is the one inducing the Frobenius norm divided by T. Finally, the isonormal process  $(W(\tmu))_{\tmu \in \R^{T \times p}}$ is defined by:
\begin{equation}
    W(\tmu):=\frac{\Tr({{\eta}^\Top \tilde{\mu}})}{T},
\end{equation}
where $\eta \in \R^{T \times p}$ is a matrix whose rows follow a centered multivariate Gaussian distribution with covariance matrix $\mathbb{I}_p$. It is easy to show that $W(\tmu)$ satisfies \Definition{def:Isonormal}.

\Theorem{Theorem:model_selection}, which can be found as Theorem 4.18 in \cite{Massart2003},  details the model selection procedure and provides us with an oracle-type inequality for this kind of estimators. The result applies for a more general model selection procedure which allows us to deal with non-linear models. Both Theorem 1 and Theorem 2 are a direct consequence of this result.

\begin{customthm}{3}{\label{Theorem:model_selection}}
Let $\{S_m\}_{m \in M}$ be some finite or countable collection of closed convex subsets of  $\Hil$. It is assumed that for any $m \in M$, there exits some almost surely continuous version $W$ of the isonormal process on $S_m$. Assume furthermore the existence of some positive and non-decreasing continuous function $\phi_m$ defined on $(0, + \infty)$ such that $\phi_m(x)/x$ is non-increasing and 
\begin{equation}{\label{ineq:gaussian_process}}
    2 \Exp\left[ \sup_{g \in S_m} \left(  \frac{W(g)-W(h)}{\norm{g-h}^2 + x^2} \right)\right] \leq x^{-2} \phi_m(x)
\end{equation}
for any positive x and any point $h$ in $S_m$. Let define $D_m>0$ such that

\begin{equation}{\label{ineq:penalization_penm}}
    \phi_m( \epsilon \sqrt{D_m})= \epsilon D_m,
\end{equation}

and consider some family of weights $\{ x_m \}_{m \in M}$ such that 

\begin{equation}{\label{ineq:sum_weights}}
    \sum_{m \in M} e^{-x_m} = \Sigma < \infty.
\end{equation}

Let $K$ be some constant with $K>1$ and take 

\begin{equation}{\label{ineq:penalty_theorem_1}}
 pen(m) \geq K \epsilon^2 \left( \sqrt{D_m} + \sqrt{2 x_m} \right)^2.
\end{equation}

Set for all $g \in H$, $\gamma (g)=\norm{g}^2 - 2 Y_{\epsilon}(g)$ and consider some collection of $p_m-$approximate penalized least squares estimators  $\{\hat{f}_m\}_{m \in M}$ i.e, for any $m \in M$,

\begin{equation}
   \gamma \left( \hat{f}_m \right) \leq \gamma (g) + \rho, \text{ for all } g\in S_m.
\end{equation}

Defining a penalized $\rho-$LSE as $\hat{f}=\hat{f}_{\hat{m}}$, the following risk bounds holds for all $f \in \Hil$

\begin{equation}
    \Exp \left[ \norm{\hat{f}-f}^2 \right] \leq C(K) \left[  \inf_{m \in M} \left( d(f,S_m)^2 + pen(m) \right) + \epsilon ( \Sigma+1) + \rho \right].
\end{equation}
\end{customthm}

\Theorem{Theorem:model_selection} requires us to have a predefined list of estimators that will be related with a list of closed convex subsets of $\Hil$. It states that we are able to recover a penalization term $pen(m)$ which allows us to find a model satisfying an oracle kind inequality if we manage to control a kind of standardized version of the isonormal process and to design a set of weights for the elements in our list of candidate models. 
\Theorem{Theorem:Massart} is a restricted version of \Theorem{Theorem:model_selection} which is more handy when dealing with the $\ell_1$-penalization term. This version of the theorem appears as Theorem A.1 in \cite{Massart2011}.

\begin{customthm}{4}{\label{Theorem:Massart}}

Let $\{S_m\}_{m \in M}$ be a countable collection of convex and compact subsets of a Hilbert space $\Hil$: lets define for any $m \in M$,
\begin{equation}
\Delta_m=\Exp\left[{\sup_{h \in S_m} W(h)}\right],
\end{equation}

and consider weights $\{x_m\}_{m \in M}$ such that 

$$
\Sigma:= \sum_{m \in M} e^{-x_m}< \infty.
$$

Let $K>1$ and assume that, for any $m \in M$,
\begin{equation}{\label{eq:pen_m}}
pen(m) \geq 2K \epsilon \left(\Delta_m + \epsilon x_m + \sqrt{\Delta_m \epsilon x_m}\right).
\end{equation}

Given a non-negative $\rho_m,m \in M$,define a $\rho_m$-approximate penalized least squares estimator as any $\hat{f} \in S_{\hat{m}},\hat{m} \in M$, such that 
\begin{equation}
\gamma(\hat{f})+pen(\hat{m}) \leq \inf_{m \in M} \Big( \inf_{h \in S_m} \gamma(h) + pen(m)+ \rho_m\Big).
\end{equation}
Then, there is a positive constant C(K) such that for all $f \in \Hil$ and $z>0$, with probability larger than $1-\Sigma e^{-z}$,
\begin{equation}
\begin{aligned}
\!\!\!\!\!\!\!\!\!\!\!\!\!\!\norm{f-\hat{f}}^2+pen(\hat{m}) \leq  C(K) \left[ \inf_{m \in M} \Big( \inf_{h \in S_m} \norm{f-h}^2 + pen(m) + \rho_m\Big) +(1+z) \epsilon^2 \right]\!\!.
\end{aligned}
\end{equation}
After integrating the inequality with respect to $z$ leads to the following risk bound: 
\begin{equation}{\label{eq:oracle_inequality}}
\begin{aligned}
\Exp \left[\norm{f-\hat{f}}^2+pen(\hat{m})\right] & \leq  C(K)\! \left[ \inf_{m \in M} \Big( \inf_{h \in S_m} \norm{f-h}^2 +pen(m)  + \rho_m\Big) +   (1+\Sigma) \epsilon^2\right]\!.
\end{aligned}
\end{equation}
\end{customthm}

Finally, we will make use of the following lemma that can be found as Lemma 2.3 in \cite{Massart2011}, a concentration inequality for real-valued random variables. 

\begin{customlem}{2}{\label{lem:inequality}}

Let $\{Z_i, i \in I\}$ be a finite family of real-valued random variables. Let $\psi$ be some convex and continuously differentiable function on $[0,b)$, with $0 < b \leq \infty$, such that $\psi(0)=\psi'(0)=0$. Assume that $\forall \gamma \in (0,b)$ and $\forall i \in I$, $\psi_{Z_i}(\gamma) \leq \psi(\gamma)$. Then, using any measurable set $B$ with $\Prob[B>0]$ we have:
\begin{equation*}
\frac{\Exp[\sup_{i \in I} Z_i \one_B] }{\Prob[B]} \leq \psi^{*-1} \left(\log \frac{|I|}{\Prob[B]}\right)\!.
\end{equation*}

In particular, if one assumes that for some non-negative number $\epsilon$, $\psi({\gamma})= \frac{\gamma^2 \epsilon^2}{2} \forall \gamma \in (0,\infty)$, then:
\begin{equation}
\frac{\Exp\left[\sup_{i \in I} Z_i \one_B \right] }{\Prob\left[B\right]} 
 \leq \epsilon \sqrt{ 2 \log\frac{|I|}{\Prob(B)}} \leq \epsilon \sqrt{2 \log|I|}+ \epsilon \sqrt{2 \log\frac{1}{\Prob(B)}}.
\end{equation}
\end{customlem}

\inlinetitle{Proof of Theorem 1}{.} Let us define the set $S_{(m,\tau)}$:
\begin{equation}{\label{expr:lasso_set}}
    S_{(m,\tau)}:=\left\{\tmu \in F_{\tau}, \norm{\tmu}_{[\tau]} \leq m \epsilon \right\},
\end{equation}
where $\norm{\tmu}_{[\tau]}=\frac{\sum_{l=1}^{d_{\tau}} I_{\tau_l} \norm{\tmu_{\tau_l}}_1}{T} $.

And $M:= \mathbb{N}^* \times \Tau$, where $\Tau$ is the set of all possible segmentations of a stream of length $T$.

We denote by  $\hat{\tau}$ and $\cpmu$
the estimators obtained by solving the Problem of Eq 5 of the main text and we will define $d_{\hat{\tau}}:=|\hat{\tau}|-1$
. Denote by $\hat{m}$ the smallest integer such that $\cpmu$ belongs to $S_{\hat{m}}$, \ie
\begin{equation}
\hat{m}=\left\lceil\frac{\mynorm\cpmu\mynorm_{[\tau]}}{\epsilon}\right\rceil\!,
\end{equation}
then,
\begin{equation*}
\begin{aligned}
\gamma{(\cpmu)} +\lambda \hat{m} \epsilon +pen(d_{\hat{\tau}}) & \leq \gamma{(\cpmu)} + \lambda \mynorm\cpmu\mynorm_{[\hat{\tau}]} + \lambda \epsilon + pen(d_{\hat{\tau}}) \\ & \leq \inf_{\tau \in \Tau} \inf_{\tmu \in S_{(m,\tau)}} \left[ \gamma{(\tmu)} + \lambda \cpnorm{\tmu} + pen(d_{\tau}) \right] + \lambda \epsilon \ \ \ \ (\text{\tiny{Definition of $\cpmu$ and $\hat{\tau}$ }})\\
 & \leq \inf_{(m,\tau) \in M} \inf_{\tmu \in S_{(m,\tau)}} \left[\gamma{(\tmu)} + \lambda m \epsilon + pen(d_{\tau}) \right] + \lambda \epsilon.
\end{aligned}
\end{equation*}
In conclusion, we have the following result: 
\begin{equation}{\label{ineq:p_approximate}}
\gamma{(\cpmu)}+pen(\hat{m},\hat{\tau}) \leq \inf_{(m,\tau) \in M}\left[\inf_{\hat{\mu} \in S_{(m,\tau)}} \gamma(\tmu) + pen(m,\tau)+ \rho\right],
\end{equation}
where $\rho=\lambda\epsilon>0$ and $pen(m,\tau)= \lambda m \epsilon + pen(d_{\tau}) >0$. 

\Ineq{ineq:p_approximate} implies $\cpmu$ is a $\rho$-approximated least squares estimator. Then, the only hypothesis that remains to be proved is Expression \ref{eq:pen_m}.

We start by getting an upper bound for $\Delta_m$. By the definition of the isonormal process $(W(\tmu))_{\tmu \in \R^{T \times p}}$, we know it is continuous. This implies that it achieves its maximum at $S_{(m,\tau)}$, a compact set, let call $\hat{g}$ this point, then:
%
\begin{comment}
\begin{equation}{\label{eq:W(h)}}
\begin{aligned}
|W(\tmu)|= \left|\frac{\tr{\zeta^{\Top}\tmu}}{T}\right|= & \left| \sum_{l=1}^{D_{\tau}} \sum_{t={\tau_{l-1}+1}}^{\tau_l} \frac{\sum_{i=i}^{p} \theta^{(i)}_{\tau_l} \dott{u_i}{\zeta_i}}{T}\right| \\ \leq &  \max_{\{i=1,..,p\}} |\dott{u_i}{\zeta_i}| \sum_{l=1}^{D_{\tau}} \frac{I_{\tau_l}}{T}\norm{\theta_{\tau_l}}_1 \\ 
\leq &  \max_{\{i=1,..,p\}} \{(\dott{u_i}{\zeta_i}) \vee (-<u_i,\zeta_i>)\} \sum_{l=1}^{D_{\tau}} \frac{I_{\tau_l}}{T}\norm{\theta_{\tau_l}}_1 \\
\leq & \sqrt{2\log 2p} \frac{\norm{\tmu}_{\tau} \epsilon}{T} \ \ \ \ \text{\tiny{\Lemma{lem:inequality}}}\\
\leq & \frac{m \epsilon}{T} \sqrt{2\log 2p} \\
\leq & \sqrt{2} \frac{m \epsilon}{T} (\sqrt{\log{p}}+\sqrt{\log{2}}). \\
\end{aligned} 
\end{equation}
\end{comment}
%
\begin{equation}{\label{eq:W(h)}}
\begin{aligned}
\Exp[|W(\hat{g})|]=\ \  \Exp\left[\left|\frac{\tr{\zeta^{\Top}\hat{g}}}{T}\right|\right]= & \Exp\left[ \left| \sum_{i=1}^{p} \sum_{t=1}^{T} \frac{\zeta^{(i)}_t \hat{g}^{(i)}_t}{T}\right| \right]\\ \leq &   \sum_{t=1}^{T} \sum_{i=1}^{p} \left| \frac{\hat{g}^{(i)}_t}{T} \right| \Exp\left[\max_{\{i=1,..,p\}} |\zeta^{(i)}_t| \right]
\\  \leq &   \sum_{l=1}^{D_{\tau}} \frac{I_{\tau_l}}{T}\norm{\hat{g}_{\tau_l}}_1 \Exp\left[ \max_{ \{i=1,..,p\}} \{ \zeta^{(i)}_t,-\zeta^{(i)}_t\} \right] \\
\leq &  \norm{\hat{g}}_{[\tau]} \sqrt{2 \log{2p}} \ \ \ \ (\text{\tiny{\Lemma{lem:inequality}}})
\\
\leq & \sqrt{2} m \epsilon \sqrt{\log{2}+\log{p}}. \ \ \ \ (\text{\tiny{\Eq{expr:lasso_set}}})
\end{aligned} 
\end{equation}

Let us define the $x_{(m,\tau)}= \gamma m + d_{\tau}L(d_{\tau})$, where $\gamma>0$. 
$L(d_{\tau})=2+\log\frac{T}{d_{\tau}}$ is a constant that just depends on the cardinality of the segmentation induced by $\tau$. Then:
\begin{equation}{\label{eq:weights_lasso}}
\begin{aligned}
\Sigma  = &  \left(\sum_{m \in N^*} e^{-\gamma m}\right)\left(\sum_{\tau \in \Tau}e^{-d_{\tau}L(d_{\tau})}\right)  \\
= &  \left(\frac{1}{e^{\gamma}-1}\right) \left( \sum_{d=1}^{T} e^{-d L(d)} | \{\tau \in \Tau , |d_{\tau}|=d\} |\right) \\
\leq &  \left(\frac{1}{e^{\gamma}-1}\right) \left( \sum_{d=1}^{T} e^{-d L(d)} \binom{T}{d}\right) \\
\leq &  \left(\frac{1}{e^\frac{\gamma}{T}-1}\right) \left( \sum_{d=1}^{T} e^{-\frac{d L(d)}{T}} \left(\frac{eT}{d}\right)^d\right)  \\
\leq & \left(\frac{1}{e^{\gamma}-1}\right) \left( \sum_{d=1}^{T} e^{-d \left(L(d)-1- \log\frac{T}{d} \right) } \right) \\
\leq & \left(\frac{1}{e^{\gamma}-1}\right) \left(\frac{1}{e-1}\right)  < \infty.
\end{aligned}
\end{equation}

Finally, let us fix $\eta=(3\sqrt{2}-2)^{-1}>0$, $K=\frac{3 }{2+\eta}>1$, $\gamma= \frac{\sqrt{\log p + L}-\sqrt{\log p +\log 2}}{K}$. It is clear $\gamma>0$ since $L>\log 2$. Then by the expressions of \Eq{eq:W(h)} and \Eq{eq:weights_lasso}, and the useful inequality $2\sqrt{ab} \leq a \eta^{-1} + b\eta $, we have: 
\begin{equation}{\label{eq:ineq_th}}
\begin{aligned}
2\frac{K\epsilon}{T} \left[ \Delta_{(m,\tau)} +\epsilon x_{(m,\tau)} + 
\sqrt{\Delta_{(m,\tau)} \epsilon x_{(m,\tau)}} \right]  & \leq 2 \frac{K \epsilon}{T} \left[ \left(1+\frac{\eta}{2}\right) \Delta_{(m,\tau)} + \left(1+\frac{\eta^{-1}}{2}\right) x_{(m,\tau)} \epsilon \right] \\
& \leq 2 \frac{K \epsilon^2}{T} \left[ \left(1+\frac{\eta}{2}\right)\left( \sqrt{2} m (\sqrt{\log{p}+\log{2}} )\right) \right. + \\  & \left. \left(1+\frac{\eta^{-1}}{2}\right) \left(\gamma m + d_{\tau}L(d_{\tau})\right) \right] \\
& \leq 3 \sqrt{2} \frac{\epsilon^2}{T} \left[ \left(\sqrt{\log p + \log 2} + K \gamma\right)m + d_{\tau}L(d_{\tau}) \right] \\
& \leq 3 \sqrt{2} \frac{\epsilon^2}{T} \left[ \left(\sqrt{\log p+ L}\right)m+ d_{\tau}L(d_{\tau}) \right] \\
&  \leq 3 \sqrt{2} \frac{\epsilon^2}{T}\left(\sqrt{\log p+ L}\right)m + \frac{d_{\tau}}{T}\left(c_1+c_2 \log\frac{T}{d_{\tau}}\right) \\
& \leq \lambda m \epsilon + pen(d_{\tau})= pen(m,\tau). 
\end{aligned}
\end{equation}

Then \Eq{eq:pen_m} is satisfied. 

We can conclude by \Eq{ineq:p_approximate}, \ref{eq:weights_lasso} and \ref{eq:ineq_th} that, if the hypotheses of \Theorem{Theorem:Massart} are satisfied, then there exists a positive constant $C(K)$ such that $\mu^* \in \R^{T \times p}$ and $z>0$, with probability larger that $1-\Sigma e^{-z}$, 
\begin{equation}
\begin{aligned}
    \frac{\norm{\cpmu-\mu^*}_F^2}{T}+pen(\hat{m})+pen(d_{\hat{\tau}})
    &  \leq C(K) \left[ \inf_{(\tau,m) \in M} \inf_{\tmu \in S_{(m,\tau)}} \left( \frac{\norm{\tmu-\mu^*}_F^2}{T} + \lambda m \epsilon + pen(d_{\tau}) \right) + \lambda\epsilon + (1+z)\epsilon^2 \right] \\
    &  \leq C(K) \left[ \inf_{\tau \in \Tau } \inf_{\tmu \in F_{\tau}} \left( \frac{\norm{\tmu-\mu^*}_F^2}{T} + \lambda \norm{\tmu}_{[\tau]} + pen(d_{\tau}) \right) + 2 \lambda \epsilon + (1+z)\epsilon^2 \right].
\end{aligned}
\end{equation}

Thanks to the last expression, we have that:
\begin{equation}
\begin{aligned}
    \frac{\norm{\cpmu-\mu^*}_F^2}{T} + \lambda \norm{\cpmu}_{[\tau]}+pen(d_{\hat{\tau}})
    & \leq C(K)  \left[ \inf_{\tau \in \Tau } \inf_{\tmu \in F_{\tau}} \left( \frac{\norm{\tmu-\mu^*}_F^2}{T} + \lambda \norm{\tmu}_{[\tau]}+ pen(d_{\tau}) \right) + 2 \lambda \epsilon + (1+z)\epsilon^2 \right]. \\
\end{aligned}
\end{equation}
After integrating this inequality, we get the desired result.

\inlinetitle{Proof of Theorem 2}{.} 
We will call $S_{D_m}$ the space generated by $m$ specific elements of the standard basis of $\R^p$ and let us define the set $S_{(D_m,\tau)}$ as:
\begin{equation}
S_{(D_m,\tau)}:=\left\{ \tmu \in F_{\tau}| \tmu_{\tau_l} \in S_{D_m} \text{ for all } l \in \{1,...,d_{\tau}\}  \right\},
\end{equation}
This implies that we restrict the means defined in each of the segments to be elements of $S_{D_m}$. 

Let define $M \subset \{1,...,p\} \times \Tau$ and let us denote $\LSEtau$ and $\LSEmu$ the solutions to the following optimization problem: 
\begin{equation}{\label{Eq:optimization_problem_2}}
\begin{aligned}
(\LSEtau,\LSEmu):= & \argmin_{ (\tau \in \Tau,\tmu \in S_{(D_m,\tau)})} \left\{
\sum_{l=1}^{d_{\tau}} \left( \sum_{t=\tau_{l-1}+1}^{\tau_{l}} \sum_{i=1}^{p}  {\frac{(\tilde{y}^{(i)}_t-\tmu^{(i)}_{\tau_l})^2}{T}}\right)+K_1\frac{D_{m}}{T} \right. \\
& \left. \qquad\qquad\qquad\quad+\frac{d_{\tau}}{T}\left(K_2+K_3 \log\frac{T}{d_{\tau}}\right) \right\}
\end{aligned}
\end{equation}

In order to obtain a oracle inequality for this estimator, we will rely on the result stated in \Theorem{Theorem:model_selection}. This means that we need to verify \Ineq{ineq:gaussian_process} and \Ineq{ineq:penalty_theorem_1} for a set of weights satisfying \Ineq{ineq:sum_weights}. 
We will begin by proving \Ineq{ineq:gaussian_process}. Let $\hat{g},\hat{f} \in S_{(D_m,\tau)}$, then we have: 
\begin{equation}{\label{ineq:dif_gp}}
\begin{aligned}
W(\hat{g})-W(\hat{h}) & =  \frac{\Tr({{\eta}^\Top \hat{g}})}{T} - \frac{\Tr({{\eta}^\Top \hat{h}})}{T} \\
&  \leq \sum_{i \in Supp_m} \sum_{t=1}^{T}  \frac{\zeta^{(i)}_t (\hat{g}^{(i)}_t-\hat{h}^{(i)}_t)}{T} \\
& \leq \sum_{i \in Supp_m} \sqrt{\sum_{t=1}^{T} \frac{(\zeta^{(i)}_t)^2}{T}} \sqrt{ \sum_{t=1}^{T} \frac{(\hat{g}^{(i)}_t-\hat{h}^{(i)}_t)^2}{T}} \ \ \ \  (\text{\tiny{Cauchy-Schwarz Ineq.}}) \\
& \leq \sqrt{\sum_{i \in Supp_m} \sum_{t=1}^{T} \frac{(\zeta^{(i)}_t)^2}{T}} \sqrt{ \sum_{i \in Supp_m} \sum_{t=1}^{T} \frac{(\hat{g}^{(i)}_t-\hat{h}^{(i)}_t)^2}{T}} \ \ \ \  (\text{\tiny{Cauchy-Schwarz Ineq.}}) \\
& = \norm{\hat{g}-\hat{h}}_{\Hil} \sqrt{\sum_{i \in Supp_m} \sum_{t=1}^{T} \frac{(\zeta^{(i)}_t)^2}{T}}.
\end{aligned}
\end{equation}
Thanks to this inequality and the fact that $\zeta^{(i)}_t$ follows a standard Gaussian distribution, we derive the following expression for each $h \in S_{(D_m,\tau)}$: 
\begin{equation}
    \begin{aligned}
    2 \Exp\left[ \sup_{\hat{g} \in S_{(D_m,\tau)}} \left(  \frac{W(\hat{g})-W(\hat{h})}{\norm{\hat{g}-\hat{h}}_{\Hil}^2 + x^2} \right)\right] & \leq x^{-1} \Exp\left[ \sup_{\hat{g} \in S_{(D_m,\tau)}} \left(  \frac{W(\hat{g})-W(\hat{h})}{\norm{\hat{g}-\hat{h}}_{\Hil} } \right)\right] \\
     & \leq x^{-1} \left[ \Exp\left[ \sup_{g \in S_{(D_m,\tau)}} \left(  \frac{W(\hat{g})-W(\hat{h})}{\norm{\hat{g}-\hat{h}}_{\Hil} } \right)^2\right] \right]^{1/2} \ \ \ \  (\text{\tiny{Jensen's ineq.}}) \\
     & \leq x^{-1} \left[ \Exp\left[ \sum_{i \in Supp_m} \frac{\sum_{t=1}^{T} (\zeta^{(i)}_t)^2 }{T} \right] \right]^{1/2} \\
     & = x^{-1} \sqrt{D_m}. \ \ \ \ (\text{\tiny{ $(\zeta^{(i)}_t)$ follows a standard Gaussian distribution}}).
    \end{aligned}
\end{equation}
We can conclude that \Ineq{ineq:gaussian_process} with $\phi_m(x)= x\sqrt{D_m}$, from which is straightforward to derive $D_m$.

Next, we define $x_{(m,\tau)}= \gamma D_m + d_{\tau}L(d_{\tau})$, where $\gamma>0$ and $L(d_{\tau})=2+\log\frac{T}{d_{\tau}}$, which is a constant that only depends on the cardinality of the segmentation induced by $\tau$. Then:
\begin{equation}{\label{eq:x_m}}
\begin{aligned}
\Sigma= \sum_{(m,\tau) \in M} e^{-x_{(m,\tau)}} = & \left(\sum_{m \in N^*, m \leq p} e^{-\gamma D_m}\right)\left(\sum_{\tau \in \Tau}e^{-d_{\tau}L(d_{\tau})}\right) \\
\leq & \left(\frac{1}{e^\gamma-1}\right) \left( \sum_{d=1}^{T} e^{-d L(d)} | \{\tau \in \Tau , |d_{\tau}|=d\} |\right) \\
\leq & \left(\frac{1}{e^\gamma-1}\right) \left( \sum_{d=1}^{T} e^{-d L(d)} \binom{T}{d}\right) \\
\leq & \left(\frac{1}{e^\gamma-1}\right) \left( \sum_{d=1}^{T} e^{-d L(d)} \left(\frac{eT}{d}\right)\right)^d \\
\leq & \left(\frac{1}{e^\gamma-1}\right) \left( \sum_{d=1}^{T} e^{-d\left(L(d)-1- \log\frac{T}{d}\right)} \right) \\
\leq & \left(\frac{1}{e^\gamma-1}\right) \left(\frac{1}{e-1}\right) < \infty.
\end{aligned}
\end{equation}

Let fix $\eta > 0$, $C> 2 + \frac{2}{\eta}$, then $K=\frac{C \eta}{2(1+\eta)}>1$. And fix $0<\delta<1$ such that $\gamma=1-\delta>0$. By using the useful inequality $2\sqrt{ab} \leq a \eta^{-1} + b\eta$.

\begin{equation}
    \begin{aligned}
    \frac{K \epsilon^2}{T} \left( \sqrt{D_m} + \sqrt{2 ( \gamma D_m +d_{\tau} L(d_{\tau})} \right)^2 & \leq  \frac{K \epsilon^2}{T}  \left( \sqrt{(1+\gamma) D_m}+ \sqrt{2 d_{\tau} L(d_{\tau})} \right)^2  \ \ \ \ (\text{\tiny{Triangle inequality}})\\
    & \leq  \frac{K \epsilon^2}{T}  \left( (1+\gamma) D_m + 2 \sqrt{2 (1+\gamma) D_m d_{\tau} L(d_{\tau}) } \right. \\
    & \left. + 2 d_{\tau} L(d_{\tau}) \right) \\
    & \leq \frac{K \epsilon^2}{T}  \left(
    (1+\gamma)D_m+ 2d_{\tau} L(d_{\tau}) \right.  \\ & \left. + (1+\gamma)D_m \eta + 2d_{\tau} L(d_{\tau}) \eta^{-1} \right) \\
    & \leq \frac{K \epsilon^2}{T} \left( (1+\gamma)(1+\eta)D_m + (2+ \eta^{-1}) d_{\tau} L(d_{\tau}) \right) \\
   & \leq  \left( C\eta(2-\delta)  \epsilon^2 \frac{D_m}{T} + C \epsilon^2  \frac{d_{\tau}}{T} L(d_{\tau})  \right) \\
   & = K_1 \frac{D_m}{T} + \frac{d_{\tau}}{T}\left(c_1+c_2 \log\frac{T}{d_{\tau}}\right) \\
    & = pen(m,\tau).
    \end{aligned}
\end{equation}

As the hypotheses of \Theorem{Theorem:model_selection} are satisfied, we obtain the desired result.

\end{document}

%% file: definitions.tex
\usepackage{graphics}
\usepackage{trimclip} % to clip around image, after scaling the image itself

\usepackage[linesnumbered,ruled,vlined]{algorithm2e}					  
\usepackage{enumitem}
\usepackage{float}
\usepackage{amsthm}
\newtheorem{theorem}{Theorem}
\newtheorem{innercustomthm}{Theorem}
\newenvironment{customthm}[1]
  {\renewcommand\theinnercustomthm{#1}\innercustomthm}
  {\endinnercustomthm}
\newtheorem{lemma}{Lemma}
\newtheorem{innercustomlem}{Lemma}
\newenvironment{customlem}[1]
  {\renewcommand\theinnercustomlem{#1}\innercustomlem{}}
  {\endinnercustomlem}
\newtheorem{definition}{Definition}
\newtheorem{innercustomdef}{Definition}
\newenvironment{customdef}[1]
  {\renewcommand\theinnercustomdef{#1}\innercustomdef{}}
  {\endinnercustomdef}
\newtheorem{Prop}{Property}

\usepackage{verbatim}
\usepackage{multirow}

\usepackage{xr}
\usepackage{balance}

\usepackage{hyperref}
\usepackage{url}
\usepackage{amsmath}
\usepackage{amssymb}

\usepackage{mathtools}

\usepackage{dsfont}
\usepackage{upgreek}
\usepackage{bbm}			% mathbb vs mathds

\usepackage{graphicx}
\usepackage{epstopdf}
\graphicspath{ {./figures/} } % setups path to find graphics and images
\usepackage{tikz}
\usepackage{float}

\usepackage{booktabs} % to thicken table lines (\toprule, \midrule, \bottomrule)

\usepackage[labelfont={small,bf}, textfont={small}]{caption}

\usepackage{subfigure}

\usepackage{xspace}

\usepackage[scaled=.7]{beramono}

\DeclareRobustCommand\sampleline[1]{%
  \tikz\draw[#1] (0,0) (0,\the\dimexpr\fontdimen22\textfont2\relax)
  -- (1.4em,\the\dimexpr\fontdimen22\textfont2\relax);%
}
\makeatletter
\g@addto@macro\bfseries{\boldmath}
\makeatother

\newlength{\phaserulewidth}
\newcommand{\setphaserulewidth}{\setlength{\phaserulewidth}}

\makeatother
\setphaserulewidth{.3pt}

\newcommand{\Sec}[1]		{Sec.\,\ref{#1}}
\newcommand{\Fig}[1]		{Fig.\,\ref{#1}}

\newcommand{\Eq}[1]			{Eq.\,\ref{#1}}

\newcommand{\Ineq}[1]			{Ineq.\,\ref{#1}}
\newcommand{\Tab}[1]		{Tab.\,\ref{#1}}
\newcommand{\Alg}[1]		{Alg.\,\ref{#1}}
\newcommand{\Theorem}[1]{Theorem\,\ref{#1}}

\newcommand{\Property}[1]{Property\,\ref{#1}}

\newcommand{\Lemma}[1]{Lemma\,\ref{#1}}
\newcommand{\Definition}[1]{Definition\,\ref{#1}}

	\newcommand{\Algorithm}[1]{Alg.\,\ref{#1}}		\newcommand{\Problem}[1]{Problem\,\ref{#1}}	
\newcommand{\ie}   			{i.e.\@\xspace}
\newcommand{\eg}   			{e.g.\@\xspace}
\newcommand{\etc}   		{etc.\xspace}
\newcommand{\wrt}   		{w.r.t.\@\xspace}

\newcommand{\one}       {\mathbf{1}}     % or  {\mathbb{1}}

\newcommand{\Exp}    {\mathbb{E}}
\newcommand{\diag}    {\operatorname{diag}}
\newcommand{\real}      {\mathbb{R}}

\newcommand{\Prob}      {\mathbb{P}}

\newcommand{\Hil}      {\mathbb{H}}
\newcommand{\transpose}		{{\mathsmaller \top}}

\newcommand{\inlinetitle}[2]  {\vspace{4pt}\noindent\textbf{\emph{#1}{#2}}}

\newcommand{\Top} {{\mathsmaller \top}}%{{\mkern-1.5mu\mathsf{T}}}

\newcommand{\Expec}[1]{\mathbb{E}[#1]}
\newcommand{\norm}[1]{\left\lVert#1\right\rVert}
\newcommand{\tr}[1]{\operatorname{tr}(#1)}
\newcommand{\cpnorm}[1]{\left\lVert#1\right\rVert_{[\tau]}}
\newcommand{\cpmu}{\hat{\tilde{\mu}}_{\hat{\tau}}}
\newcommand{\tmu}{\tilde{\mu}}
\newcommand{\dott}[2]{\langle #1,#2 \rangle}

\newcommand{\Tau}{\mathcal{T}}
\newcommand{\Taud}{\mathcal{T}^d}
\DeclareMathOperator{\Tr}{tr}
\newcommand{\mynorm}{|\!|}

\newcommand{\LSEmu}{\hat{\tau}^{\text{LSE}}}
\newcommand{\LSEd}{\hat{d}^{\text{LSE}}}
\newcommand{\LSEtau}{\hat{\tilde{\mu}}^{\text{LSE}}_{\hat{\tau}^{\text{LSE}}}}

\DeclareMathOperator*{\argmin}{\arg\!\min}

\DeclareMathOperator*{\argminA}{arg\,min}
\DeclareMathOperator*{\R}{\mathbb{R}}

\newcounter{marginNoteCounter}